\documentclass[oneside]{article}
\usepackage{xcolor}
\usepackage{ecj,palatino,epsfig,latexsym,natbib}
\usepackage{mathrsfs}
\usepackage{amsfonts}
\usepackage{mathtools}
\usepackage{amsmath}
\usepackage{bbm}
\usepackage{amsthm}
\usepackage{tikz}
\usetikzlibrary{shapes,arrows,fit}
\usetikzlibrary{positioning}
\usetikzlibrary{decorations.pathreplacing}
\usetikzlibrary{calc}
\usepackage{pgfplots}
\usepackage{subcaption}
\usepackage{pgfplotstable}
\usepackage{mathtools}
\usepackage[scleft,ruled]{algorithm2e}
\usepackage[a4paper]{geometry}
\usepackage{float}
\usepackage{hyperref}

\pgfplotsset{compat=1.16}
\newcommand{\graphScale}{0.7}

\newtheorem{property}{Property}
\DeclareMathOperator*{\argmax}{arg\,max}

\newtheorem{theorem}{Theorem}

\newcommand{\pluseq}{\mathrel{+}=}

\newcommand{\Norm}[1]{\left\lVert#1\right\rVert}

\begin{document}

\ecjHeader{x}{x}{xxx-xxx}{2021}{}{N.H. Hamilton, E.W. Fulp}
\title{\bf Evolutionary Optimization of High-Coverage Budgeted Classifiers}  

\author{\name{ Nolan H. Hamilton} \hfill \addr{haminh16@wfu.edu}\\ 
        \addr{Department of Computer Science, Wake Forest University, 
        Winston-Salem, NC, USA}
\AND
       \name{ Errin W. Fulp} \hfill \addr{fulp@wfu.edu}\\
        \addr{Department of Computer Science, Wake Forest University, 
        Winston-Salem, NC, USA}
}

\maketitle

\begin{abstract}
Classifiers are often utilized in time-constrained settings where labels must be assigned to inputs quickly. To address these scenarios, budgeted multi-stage classifiers (MSC) process inputs through a sequence of \textit{partial} feature acquisition and evaluation steps with early-exit options until a confident prediction can be made. This allows for fast evaluation that can prevent expensive, unnecessary feature acquisition in time-critical instances. However, performance of MSCs is highly sensitive to several design aspects---making optimization of these systems an important but difficult problem.

To approximate an initially intractable combinatorial problem, current approaches to MSC configuration rely on well-behaved surrogate loss functions accounting for two primary objectives (processing time, error). These approaches have proven useful in many scenarios but are limited by analytic constraints (convexity, smoothness, etc.) and do not manage additional performance objectives. Notably, such methods do not explicitly account for an important aspect of real-time detection systems---the ratio of ``accepted'' predictions satisfying some confidence criterion imposed by a risk-averse monitor.

This paper proposes a problem-specific genetic algorithm, \texttt{EMSCO}, that incorporates a terminal reject option for indecisive predictions and treats MSC design as an evolutionary optimization problem with distinct objectives (accuracy, processing cost, coverage). The algorithm's design emphasizes Pareto efficiency while respecting a notion of aggregated performance via a unique scalarization. Experiments are conducted to demonstrate \texttt{EMSCO}'s ability to find global optima in a variety of $\Theta(k^n)$ solution spaces, and multiple experiments show \texttt{EMSCO} offers key advantages compared to alternative budgeted approaches.
\end{abstract}


\section{Introduction}\label{sec:introduction}
    Machine learning techniques have become popular among practitioners for deployment in real-time detection scenarios. However, traditional out-of-box methods are not always appropriate for resource-constrained contexts. Many application domains present implicit latency constraints between observation of inputs and label prediction, and a common
    difficulty regards balancing processing time/cost with accuracy \citep{ji2007,trapeznikov2013,xu2012}. 

    In these \textit{budgeted} learning settings, the amount of resources (e.g., time) expended while evaluating inputs represents the \textit{cost of classification}. This cost is commonly specified further as the sum of \textit{feature acquisition cost} and \textit{classifier evaluation cost}. Feature acquisition refers to the process of computing informative metrics for classification that are not present during the initial observation of inputs; Classifier evaluation cost is a measure of the resources spent evaluating an input with a classifier that has been trained on the acquired features. Note, cost of classification is often dominated by feature acquisition cost \citep{kusner,xu2012}.

    For example, consider the task of classifying e-mail messages as spam. A large influx of messages must be evaluated quickly, and informative features such as number of URLs, subject-line character diversity, sender reputation score, etc. require CPU-time to compute. A classification scheme that requires a large set of such features to label each input may therefore be impractical depending on the number of messages and desired processing speed. Fortunately, it is often possible to use smaller subsets of features to correctly classify a fraction of test-time inputs ~\citep{trapeznikov2013,xu2012,xu2014,ji2007,nan16}.
    
    Multi-stage classifiers leverage this notion that some inputs may require only a few inexpensive features to be confidently labeled, while other ``harder'' inputs may require additional features for discernment. Using this design, a set of classifiers with growing feature sets are arranged in sequence to evaluate inputs~\citep{trapeznikov2013,ji2007,wang2014}. Once an input receives a confident class prediction, processing ceases and no cost is incurred for the remaining features that were not used.
    Specific use-cases of sequential, budgeted learning can be found across a diverse range of application domains \citep{dundar07,hamilton20202,nogueira2019}. 
    
    While budgeted learning schemes present an opportunity for efficient processing of inputs in time-critical instances, there are a multitude of important design considerations that can affect performance of these protocols, making configuration a difficult optimization problem. However, given the potential benefits, design of budgeted classifiers has become an active area of research in the past decade \citep{trapeznikov2013, ji2007, xu2014, xu2012, nan16, peter17,jansich}. 
    
    \subsection{Related Work}
    In this section, we offer a brief survey of prominent budgeted classification algorithms. Naturally, a common approach to optimizing budgeted classifiers involves constructing a surrogate loss function that penalizes error \textit{and} cost of classification. This well-behaved loss function is then minimized with descent techniques. 
    
    \citep{ji2007} treats the problem as a partially-observable Markov process and requires estimation of class probability models that restrict applicability in some use cases.   \citep{trapeznikov2013} formulates the problem with reject options at each stage comprised of weak learners. These reject classifiers are optimized using coordinate descent on a feature-budgeted empirical risk function. To avoid difficult combinatorial aspects, the order of features is fixed beforehand. The result is a solution with guaranteed local optimality---but only with respect to the chosen ordering of features and stages. \citep{wang2014} establishes an improvement over the myopic approach proposed in \citep{trapeznikov2013} by devising a convex loss function. The authors are then able to guarantee globally optimal solutions within the context of their formulation and a fixed feature/stage order.
 
    While many budgeted protocols rely on a fixed stage design, several existing methods incorporate feature selection/order implicitly. \citep{xu2012} proposes \textit{GreedyMiser}---a feature-budgeted variant of stage-wise regression \citep{friedman2001}. Limited-depth regression trees are used as weak learners and are constructed with a modified impurity function accounting for cost of feature extraction. These weak learners are combined to form a final classifier. For its relative simplicity and good scaling to high-dimensional data, GreedyMiser has become one of the most popular and best-cited feature-budgeted approaches to classification.  \citep{xu2014} proposes \textit{Cost-Sensitive Tree of Classifiers}; This method builds a tree of classifiers optimized for a specific subset of the input space.
The input space is partitioned based on classifier predictions. Classifiers that reside deep in the tree become ``experts'' for a small subset of the input space, and intermediate classifiers determine the path of instances through the tree. CSTC is trained with a single global loss function, where the test-time cost penalty is a relaxation of the expected cost.

    While the described methods provide a pathway to efficient classification of inputs in resource-constrained settings and significant theoretical guarantees, several limitations are consistent throughout this body of work. First, many of the methods do not optimize the order of the features/stages \citep{trapeznikov2013,wang2014} and instead resort to increasing cost heuristics to choose an ordering of features (expensive features placed at later stages). This can be inefficient if many cheap features are uninformative. Additionally, intuitively modeling complex feature interactions beforehand can prove difficult, and such interactions may be consequential if variables are more/less discriminative when grouped together \citep{ji2007}.

     Furthermore, these methods rely on a single homogeneous loss function that can fail to properly manage multiple objectives in a fitness landscape that is not well-behaved. This can be problematic in scenarios where important non-dominated solutions are not supported by scalarization and several objectives are consequential~\citep{deb01,boyd}. There are important performance aspects beyond cost and accuracy in real-time prediction systems, and one frequently neglected metric is \textit{conclusiveness}---the ratio of predictions deemed sufficiently confident to be recorded and not discarded by a terminal \textit{reject option}. This metric is also referred to as ``coverage''\footnote{In this manuscript, we use the terms ``coverage'' and ``conclusiveness'' interchangeably.} in the domain of selective classification (SC) \citep{yaniv10,geifman2017}. With this in mind, the cost/accuracy framework embraced by alternative methods is potentially limiting, and a need arises for a budgeted approach with a terminal reject option that can effectively manage several objectives explicitly.

    Note, there are several loosely related learning paradigms that will not be addressed in this manuscript. \textit{Classifier cascades}, for example, leverage an assumption of class imbalance and are restricted to binary classification \citep{viola}.

    \subsection{Evolutionary Approaches to Multi-Stage Classifier Design}
    Many of the issues discussed above can be alleviated by formulating MSC design in a way that incorporates a terminal reject option, considers order/composition of stages, and  manages multiple objectives in a Pareto efficient manner. This paper builds on a preliminary approach\footnote{In contrast to the method proposed in this paper, the preliminary approach uses generic evolutionary operators, a less robust elitism procedure, and does not provide several important algorithmic guarantees.} \citep{hamilton2020} of the authors to treat multi-stage classifier configuration as a combinatorial optimization problem to be solved with evolutionary algorithms(EAs). EAs employ a population-based approach to optimization in which the generation/population $G_1$ is a product of mutation, selection, and crossover on the solutions in $G_0$.  These algorithms possess desirable properties in the setting of multi-objective optimization \citep{deb01} when gradients are not available and mathematical flexibility is needed.
    % somewhere in the paper, we should really make tep oint that we are free to use whatever classifiers we want for stages!!

  Three objectives are considered: conclusive accuracy, feature acquisition cost, and conclusiveness (See Section \ref{objectives} for a detailed description of these objectives.) Leveraging the flexibility of EAs, \texttt{EMSCO} is unique compared to alternative budgeted approaches in that it defines these objectives with indicators. This allows for a natural treatment of the problem that is less dependent on approximation.
Additionally, by incorporating non-domination level in the fitness function and applying domain-specific knowledge in the evolutionary operators, \texttt{EMSCO} ensures several important properties described later (Section \ref{sec:fitness}).

\section{A Formal Description of Multi-Stage Classification} \label{sec:setup}

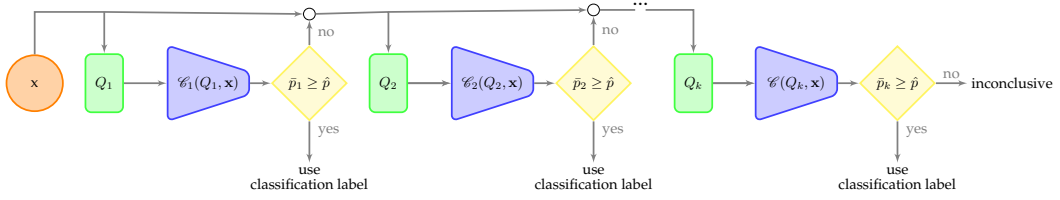
\begin{figure*}[t] 
\begin{center}
\resizebox{\textwidth}{!}
{ 
% \begin{figure}
% \begin{center}
% \resizebox{\textwidth}{!}{  
% Define block styles
\tikzstyle{decision} = [diamond, very thick, draw=yellow!75, fill=yellow!20,
    text width=4.25em, text badly centered, node distance=3.2cm, inner sep=0pt]
\tikzstyle{block} = [rectangle, very thick, draw=green!75, fill=green!20,
    text width=6em, text centered, rounded corners, minimum height=4em]
\tikzstyle{stage} = [rectangle, very thick, draw=black, fill=blue!20,
    text width=5em, text centered, rounded corners, minimum height=4em]
\tikzstyle{feature} = [rectangle, very thick, draw=green!75, fill=green!20,
    text width=2em, text centered, rounded corners, minimum height=4em]
\tikzstyle{trap} = [trapezium, draw, minimum width = 1.5cm,
trapezium left angle = 108, trapezium right angle = 108, shape border rotate = 90, very thick, draw=black!75, fill=black!20, text width=5em, text centered, rounded corners, minimum height=4em]
\tikzstyle{data} = [trapezium, draw, trapezium left angle=60,
  trapezium right angle=-60, very thick, draw=green!75, fill=green!20,
    text width=5em, text centered, rounded corners, minimum height=4em]
\tikzstyle{line} = [draw, very thick, color=black!50, -latex']
\tikzstyle{cloud} = [draw=black!75, circle, very thick, fill=black!20, node distance=2.5cm,
    minimum height=3.85em]
\tikzstyle{decision} = [diamond, very thick, draw=yellow!75, fill=yellow!20,
    text width=4.25em, text badly centered, node distance=3.2cm, inner sep=0pt]
\tikzstyle{block} = [rectangle, very thick, draw=green!75, fill=green!20,
    text width=6em, text centered, rounded corners, minimum height=4em]
\tikzstyle{stage} = [rectangle, very thick, draw=black, fill=blue!20,
    text width=5em, text centered, rounded corners, minimum height=4em]
\tikzstyle{feature} = [rectangle, very thick, draw=green!75, fill=green!20,
    text width=2em, text centered, rounded corners, minimum height=4em]
\tikzstyle{trap} = [trapezium, draw, minimum width = 1.5cm,
trapezium left angle = 108, trapezium right angle = 108, shape border rotate = 90, very thick, draw=blue!75, fill=blue!20, text width=5em, text centered, rounded corners, minimum height=4em]
\tikzstyle{data} = [trapezium, draw, trapezium left angle=60,
  trapezium right angle=-60, very thick, draw=green!75, fill=green!20,
    text width=5em, text centered, rounded corners, minimum height=4em]
\tikzstyle{line} = [draw, very thick, color=black!50, -latex']
\tikzstyle{cloud} = [draw=orange!100, circle, very thick, fill=orange!35, node distance=2.5cm,
    minimum height=3.85em]

\begin{tikzpicture}[node distance = 2.5cm and 1.0cm, auto, inner sep = 1mm, scale=.75]
   % \draw[black,very thick, help lines] (-1, -7) grid (15, 2);

   % place EA nodes
   \node [cloud] (start) {\shortstack{x}};
   \node [feature, right = 0.5 of start] (F1) {{$\Large{Q_1}$}};
   \coordinate[above = 1cm of F1] (d1);
   \node [trap, right of = F1] (stage1){$\mathscr{C}_1(Q_1,\mathbf{x})$};
%    \node [below = 1cm of stage1] (F1) {$F_{1}$};
   \node [decision, right = 0.5cm of stage1] (decision1) {$\bar{p}_{1} \geq \hat{p}$};  
   \node [draw, circle, above = 0.6cm of decision1] (continue1) {};
   \node [below = 1cm of decision1] (done1) {\shortstack{use \\ classification label}};

   \node [feature, right = 0.5cm of decision1] (F2) {{$\Large{Q_2}$}};
   \coordinate[above = 1.0cm of F2] (d2);
   \node [trap, right of = F2] (stage2) {$\mathscr{C}_2(Q_2,\mathbf{x})$};
   % \node [below = 1cm of stage2] (F2) {$F_{2}$};
   \node [decision, right = 0.5cm of stage2] (decision2) {$\bar{p}_{2} \geq \hat{p}$};  
   \node [draw, circle, above = 0.7cm of decision2] (continue2) {};
   \node [below = 1cm of decision2] (done2) {\shortstack{use \\ classification label}};

   \coordinate[right = 0.85cm of continue2] (upper_dot_start);
   \coordinate[right = 0.25cm of upper_dot_start] (upper_dot_end);
   \coordinate[right = 0.25cm of decision2] (dot_start);
   \coordinate[right = 0.25cm of dot_start] (dot_end);

   \node [feature, right = 0.5 of dot_end] (Fn) {{$\Large{Q_k}$}};
   \coordinate[above = 1.05cm of Fn] (dn);
   \node [trap, right = 1.0cm of Fn] (stagen) {$\mathscr{C}(Q_k,\mathbf{x})$};
   % \node [below = 1cm of stagen] (Fn) {$F_{n}$};
   \node [decision, right = 0.5cm of stagen] (decisionn) {$\bar{p}_{k} \geq \hat{p}$};  
   \node [below = 1cm of decisionn] (donen) {\shortstack{use \\ classification label}};
   \node[right = 0.75cm of decisionn] (idk) {\shortstack{inconclusive}};

   % draw edges
   \path [line] (start) |- (d1) -- (continue1);
   \path [line] (d1) -- (F1);
   \path [line] (F1) -- (stage1);
   \path [line] (stage1) -- (decision1);
   \path [line] (decision1) -- node[swap, midway] {~no} (continue1);

   \path [line] (F2) -- (stage2);
   \path [line] (continue1) -- (d2) -- (F2);
   \path [line] (d2) -- (continue2);
   \path [line] (decision1) -- node[near start] {~yes} (done1);
   \path [line] (stage2) -- (decision2);
   \path [line] (F2) -- (stage2);
   \path [line] (decision2) -- node[swap, midway] {~no} (continue2);
%    \path [line] (decision2) -- node[midway] {no} (dot_start);
   \path [line] (decision2) -- node[near start] {~yes} (done2);

   \path (upper_dot_start) -- node[]{\Large{...}} (upper_dot_end);
   \path (dot_start) -- (dot_end);

   \path [draw, very thick, color=black!50] (continue2) -- (upper_dot_start);

   \path [line] (Fn) -- (stagen);
   \path [line] (Fn) -- (stagen);
   \path [line] (upper_dot_end) -- (dn) -- (Fn);
   \path [line] (stagen) -- (decisionn);
   \path [line] (decisionn) -- node[midway] {no} (idk);
   \path [line] (decisionn) -- node[near start] {~yes} (donen);

\end{tikzpicture}

}
\caption{Design of a $k$-stage classifier with confidence threshold $\hat{p}$. Inputs are processed sequentially through classification stages until they achieve a class probability greater than $\hat{p}$ or all features have been exhausted. \label{fig:workflow}}

\end{center}

\end{figure*}

    This section details the setting in which \texttt{EMSCO} is deployed. See Table \ref{tab:notation} for a description of notation and symbols used throughout the paper and Figure \ref{fig:workflow} for a visual depiction of \texttt{EMSCO}'s workflow.  
    
    Assume a classification scenario in which inputs are evaluated with a potential feature set $\mathcal{F}$, where $||\mathcal{F}|| = n$. Upon initial observation, an input $\mathbf{x} \in \mathcal{X}_{test}$ begins with no acquired features, and features are thereafter attained sequentially through stages. That is, stage $j$ acquires a subset of features $Q_j \subseteq \mathcal{F}$ to evaluate\footnote{The stage $\mathscr{C}(Q_j, \mathbf{x})$ can use any classification algorithm suitable for the given context. For the sake of generalization and computational efficiency, \texttt{EMSCO} employs $\ell_2$-regularized logistic regression at each stage in this paper.} input $\mathbf{x}$.  The stage-$j$ classifier $\mathscr{C}(Q_j,\mathbf{x})$, learned on features in $Q_j$ during training-time, returns a set of prediction confidences $\mathscr{P_{\mathbf{x_j}}} = \{p_1, p_2, \ldots, p_l\}$ corresponding to $l$ possible labels for the input. Note, every feature has a corresponding cost of acquisition given by the set $\mathcal{C}$---acquiring feature $\mathcal{F}_i$ for evaluation of $\mathbf{x}$ increases the classification cost of $\mathbf{x}$ by $\mathcal{C}_i$. After a particular feature has been acquired for an input in stage $j$, it can be referenced in stage $j+1, j+2,\ldots,k$ for no cost.
    
     We assume a \textit{risk-averse monitor} has decided to accept only confident predictions. At preterminal stages $j < k$, a reject decision is made to determine whether the input prediction is sufficiently confident to be accepted or if additional processing is necessary (in which case, we say $\mathbf{x}$ has been ``rejected'' to the next stage). Let $\bar{p_j} =  \max \mathscr{P_{\mathbf{x_j}}}$ and $\hat{p}$ be a confidence threshold specified by the monitor. At stage $j<k$, if $\bar{p_j} < \hat{p}$, the input is sent to stage $j+1$ where it is evaluated with feature set $Q_{j+1}$ (Note, $Q_j \subset Q_{j+1}$). Conversely, if $\bar{p_j} \geq \hat{p}$, processing ceases, and $\mathbf{x}$ is assigned the label corresponding to the maximum class probability.
    
     At the terminal stage $k$, if $\bar{p_k} < \hat{p}$, evaluation of $\mathbf{x}$ is deemed \textit{inconclusive} with no label assigned. In this case, we say that $\mathbf{x}$ has been \textit{terminally} rejected. This is a safe but generally undesirable result as it decreases utility of the system---while desiring at least $\hat{p}$ accuracy, the above-mentioned risk-averse monitor also desires that the system yield insight. Figure \ref{fig:threshold} depicts the effect of $\hat{p}$ on system performance in a generic setting.
     
     With a confidence threshold specified \textit{a priori}, our aim is to find a stage configuration that provides high accuracy (at least $\hat{p}$), low processing cost, and high coverage. See Section \ref{problemdef} for the formal problem statement.
 
    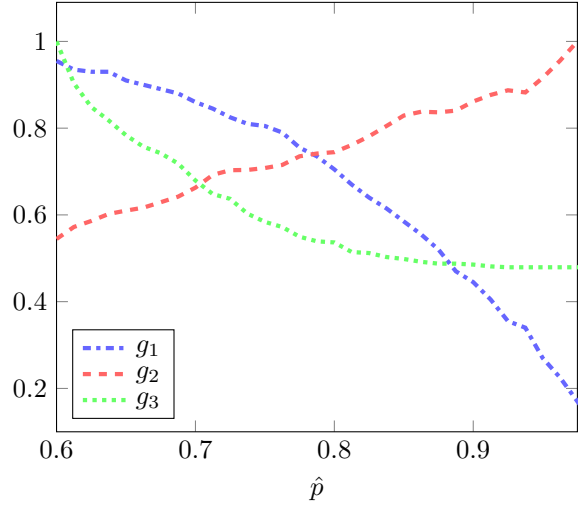
\begin{figure} \label{fig:threshold}
 \begin{center}
\begin{tikzpicture}

\begin{axis}[
    every axis plot/.append style={ultra thick},xlabel={$\hat{p}$},xmin=.6,xmax=.975,ymin=.1,legend pos=south west,title={Effect of Confidence Threshold $\hat{p}$}]

\addplot[color=blue!60, dash dot] coordinates {
(0.6, 0.955)
(0.6124999999999999, 0.935)
(0.6249999999999999, 0.9299999999999999)
(0.6374999999999998, 0.9299999999999999)
(0.6499999999999998, 0.91)
(0.6624999999999998, 0.9)
(0.6749999999999997, 0.89)
(0.6874999999999997, 0.88)
(0.6999999999999996, 0.86)
(0.7124999999999996, 0.845)
(0.7249999999999995, 0.825)
(0.7374999999999995, 0.81)
(0.7499999999999994, 0.8049999999999999)
(0.7624999999999994, 0.79)
(0.7749999999999994, 0.755)
(0.7874999999999993, 0.735)
(0.7999999999999993, 0.7050000000000001)
(0.8124999999999992, 0.6699999999999999)
(0.8249999999999992, 0.64)
(0.8374999999999991, 0.615)
(0.8499999999999991, 0.585)
(0.862499999999999, 0.5549999999999999)
(0.874999999999999, 0.52)
(0.887499999999999, 0.47)
(0.8999999999999989, 0.44499999999999995)
(0.9124999999999989, 0.405)
(0.9249999999999988, 0.355)
(0.9374999999999988, 0.33999999999999997)
(0.9499999999999987, 0.27)
(0.9624999999999987, 0.22499999999999998)
(0.9749999999999986, 0.17000000000000004)
(0.9874999999999986, 0.12)
};

\addplot[color=red!60,dashed] coordinates {
(0.6, 0.5445026178010471)
(0.6124999999999999, 0.5721925133689839)
(0.6249999999999999, 0.5860215053763441)
(0.6374999999999998, 0.6021505376344086)
(0.6499999999999998, 0.6098901098901099)
(0.6624999999999998, 0.6166666666666667)
(0.6749999999999997, 0.6292134831460674)
(0.6874999999999997, 0.6420454545454546)
(0.6999999999999996, 0.6627906976744186)
(0.7124999999999996, 0.6923076923076923)
(0.7249999999999995, 0.703030303030303)
(0.7374999999999995, 0.7037037037037037)
(0.7499999999999994, 0.7080745341614907)
(0.7624999999999994, 0.7151898734177216)
(0.7749999999999994, 0.7350993377483444)
(0.7874999999999993, 0.7414965986394558)
(0.7999999999999993, 0.7446808510638298)
(0.8124999999999992, 0.7611940298507462)
(0.8249999999999992, 0.78125)
(0.8374999999999991, 0.8048780487804879)
(0.8499999999999991, 0.8290598290598291)
(0.862499999999999, 0.8378378378378378)
(0.874999999999999, 0.8365384615384616)
(0.887499999999999, 0.8404255319148937)
(0.8999999999999989, 0.8614606741573034)
(0.9124999999999989, 0.8765432098765432)
(0.9249999999999988, 0.8873239436619719)
(0.9374999999999988, 0.8823529411764706)
(0.9499999999999987, 0.9159259259259259)
(0.9624999999999987, 0.9555555555555556)
(0.9749999999999986, 1.0)
(0.9874999999999986, 1.0)
};

\addplot[color=green!60,dotted] coordinates {
(0.6, 1.0)
(0.6124999999999999, 0.903448275862069)
(0.6249999999999999, 0.8488120950323974)
(0.6374999999999998, 0.81789802289282)
(0.6499999999999998, 0.7836490528414756)
(0.6624999999999998, 0.7572254335260116)
(0.6749999999999997, 0.7422096317280453)
(0.6874999999999997, 0.7184643510054844)
(0.6999999999999996, 0.6793431287813311)
(0.7124999999999996, 0.6479802143446002)
(0.7249999999999995, 0.6369529983792545)
(0.7374999999999995, 0.602760736196319)
(0.7499999999999994, 0.5839524517087668)
(0.7624999999999994, 0.5720524017467249)
(0.7749999999999994, 0.5496503496503496)
(0.7874999999999993, 0.5390946502057613)
(0.7999999999999993, 0.5365187713310581)
(0.8124999999999992, 0.5143979057591623)
(0.8249999999999992, 0.5120521172638437)
(0.8374999999999991, 0.5028790786948176)
(0.8499999999999991, 0.4984147114774889)
(0.862499999999999, 0.4918648310387985)
(0.874999999999999, 0.48759305210918114)
(0.887499999999999, 0.48759305210918114)
(0.8999999999999989, 0.48548486720197653)
(0.9124999999999989, 0.48132271892222905)
(0.9249999999999988, 0.4792682926829268)
(0.9374999999999988, 0.4792682926829268)
(0.9499999999999987, 0.4792682926829268)
(0.9624999999999987, 0.4792682926829268)
(0.9749999999999986, 0.4792682926829268)
(0.9874999999999986, 0.4792682926829268)
};
\legend{$g_1$,$g_2$,$g_3$}
\end{axis}
\end{tikzpicture}
\caption{\small{Example system behavior under varying $\hat{p}$. Generally, higher $\hat{p}$ corresponds to greater conclusive accuracy but lower coverage and increased cost. Objectives were measured using a cost-ordered two-stage model. We used Scikit-Learn's  \citep{pedregosa2011} function \texttt{make\_classification()} to generate a synthetic data set with fifteen features, five-hundred samples, and binary labels.}}
\label{fig:threshold}
\end{center}
\end{figure}

    \begin{table} 
    \begin{small}
    \begin{center}
    \renewcommand{\arraystretch}{1.33}
    \begin{tabular}{ccl}
    \textbf{Symbol} & \shortstack{\textbf{Description}} \\ \hline \hline
    $\beta$ & stage count bias parameter \\ \hline
    $b$ & elitism proportion\\ \hline
    $\mathcal{C}$ & feature costs \\ \hline
    $\mathscr{C}(Q_j,\mathbf{x})$ & returns class probabilities for $\mathbf{x}$ after $j$\textsuperscript{th} stage evaluation \\ \hline
    $E_0^*$ & unique non-dominated solutions in a given generation \\ \hline
    $\mathscr{E}(g_1,g_2,g_3)$ & objective Euclidean norm \\ \hline
    $\mathcal{F}$ & feature set \\ \hline
    $G_h$ & refers to the $h$\textsuperscript{th} generation \\ \hline
    $\Norm{G}$ & population size \\ \hline
    $g$ & convergence parameter \\ \hline
    $k$ & max number of stages \\ \hline
    $M$ & elite population size (unfixed) \\ \hline
    $\hat{m}$ & mutation rate \\ \hline
    $n$ & number of features \\ \hline
    $N$ & number of test records \\ \hline
    $\mathscr{N}(G_h)$ & returns the non-dominated front in generation $G_h$ \\ \hline
    $\mathcal{P}_{(n,j)}$ & ordered $j$-partitions of $\mathcal{F}$ \\ \hline
    $\bar{p_j}$ & maximum class probability after $j$\textsuperscript{th} stage\\ \hline
    $\hat{p}$ & confidence threshold \\ \hline
    $\Norm{Q}$ & stage count of solution $Q$ \\ \hline
    $Q_j$ & features in stage $j$ \\ \hline
    $[Q]$ & chromosome representation of Solution $Q$ \\ \hline
    $[Q]_i$ & stage assignment for feature $i$ \\ \hline
    $\Bar{[Q]}_{max}$ & Element-wise mean vector of top-scoring chromosomes\\ \hline
    $\hat{r}$ & recombination rate \\ \hline
    $r_{max}$ & rank of the non-dominated solutions in a specified population\\ \hline
    $\mathcal{S}_{(n,k)}$ & all solutions with at most $k$ stages \\ \hline
    $T_i$ & cost class of $i$\textsuperscript{th} feature \\ \hline
    $\lceil \cdot \rceil$ & returns next smallest integer \\ \hline
    $\mathbbm{1}[\cdot]$ & indicator function \\ \hline
    $\succ$ & Pareto dominance relation \\ \hline
    \hline 
    \end{tabular}
    \label{tab:import2}  
    \caption{Notation and Symbols used in Manuscript}\label{tab:notation}
    \end{center}
    \end{small}
    \end{table}
    \subsection{Solution Space}
    To provide insight into the nature and difficulty of the problem under our formulation, we offer an analysis of the solution space. Solutions are $k$-\textit{partition}s of the feature set $\mathcal{F}$. That is, $$\bigcup_{j=1}^{j=k} Q_j = \mathcal{F}$$ where $Q_j \neq \emptyset$ denotes the new features acquired in stage $j$. The order of the stages is also considered. For example, if $\mathcal{F} = \left \{\mathcal{F}_1, \mathcal{F}_2, \mathcal{F}_3, \mathcal{F}_4 \right \} $, then
    $$Q = \left \{ \{\mathcal{F}_1, \mathcal{F}_2\}, \{\mathcal{F}_3,\mathcal{F}_4\} \right \} \neq \left \{ \{\mathcal{F}_3, \mathcal{F}_4\}, \{\mathcal{F}_1,\mathcal{F}_2\} \right \}$$
    The number of ordered $k$-partitions $||\mathcal{P}_{(n,k)}||$ on a feature set with dimension $n$ is then computed by multiplying the number of unordered partitions by $k!$:
    \begin{equation} 
     ||\mathcal{P}_{(n,k)}|| = k! \cdot S_2(n,k) = \sum\limits_{i=0}^{k}(-1)^{k-i}{k \choose i}i^n,
    \end{equation}
    where $S_2(n,k)$ represents the number of unordered $k$-partitions of a set with $n$ elements, also known as a \textit{Stirling number of the second kind} \citep{knuth1998}. In practice, \texttt{EMSCO} treats $k$ as an upper bound for stage count---the search space, $\mathcal{S}_{(n,k)}$, is the set of all solutions with \textit{up to and including} $k$ stages. That is, 
    \begin{equation} \label{solutionspace}
    \mathcal{S}_{(n,k)} = \bigcup_{j=1}^{j=k}  \mathcal{P}_{(n,j)}.
    \end{equation}
    The following theorem addresses the limiting behaviour of  $||\mathcal{S}_{(n,k)}||$.
    \begin{theorem} \label{thm:solutionspace}
     For a fixed $k$ such that $k \leq n$, the number of candidate solutions $||\mathcal{S}_{(n,k)}||$ is asymptotically equivalent to $k^n$.
    \begin{proof}
    We first show that the ratio of quantities $||\mathcal{P}_{(n,k)}||$ and $k^n$ approaches one as $n \rightarrow \infty$. Since $k$ is fixed, every term in right-hand-side of the expression below approaches zero except the last.
    $$\lim_{n\to\infty} \frac{1}{k^n}\sum\limits_{i=0}^{k}(-1)^{k-i}{k \choose i}i^n = \lim_{n\to\infty} \pm \frac{\binom{k}{0}0^n}{k^n} \pm \frac{\binom{k}{1}1^n}{k^n} \pm \ldots + \frac{\binom{k}{k}k^n}{k^n},$$
    where $\frac{\binom{k}{k}k^n}{k^n} = 1$. Thus, $||\mathcal{P}_{(n,k)}|| \sim k^n$. Since there is no intersection between terms in the union on the RHS of Equation \ref{solutionspace}, $$\sum_{j=1}^{j=k} ||\mathcal{P}_{(n,j)}|| = ||\mathcal{S}_{(n,k)}||.$$ 
    Using this fact and $||\mathcal{P}_{(n,j)}|| \sim j^n$, we can quickly show:
    $$\lim_{n\to\infty}  ||\mathcal{S}_{(n,k)}|| =  \lim_{n\to\infty} \frac{1}{k^n}\sum_{j=1}^{j=k}  ||\mathcal{P}_{(n,j)}|| =  \lim_{n\to\infty} \frac{1}{k^n}\sum_{j=1}^{j=k} j^n = \lim_{n\to\infty} \left(\frac{k}{k}\right)^n = 1.$$ That is, $||\mathcal{S}_{(n,k)}|| \sim k^n$.
    \end{proof}
    \end{theorem}
    In practice, all features in stage $j$ are automatically appended to stage $j+1$ as they are already acquired and can be referenced for no cost. However, this technicality does not affect the above analysis, since it depends only on the unique elements acquired at each stage.

 As our defining structure for MSCs, ordered feature set partitions possess several beneficial properties. In many cases, certain features are uninformative alone but become highly discriminative when evaluated together \citep{ji2007}. By allowing stages to contain multiple features, these scenarios are acknowledged implicitly during optimization. Additionally, using the representation proposed in Section \ref{representation}, ordered feature set partitions can be conveniently modeled as chromosomes.
 
 Notice that $\hat{p}$ is fixed and does not affect solution space $\mathcal{S}_{(n,k)}$ since it is assumed this value is set \textit{a priori} by a risk-averse monitor whose misclassification penalties and resource constraints warrant such a threshold.
    \subsection{Objectives}\label{objectives}
    
\begin{figure}
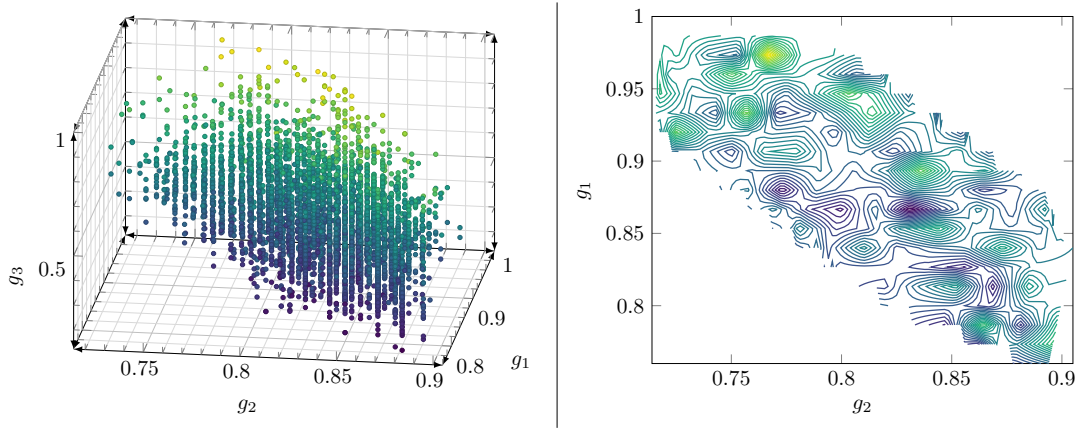

\begin{center}
\resizebox{\textwidth}{!}{% 
\begin{tabular}{c|c}
% \multicolumn{2}{c}{Sample Objective Space} \\
\usebox{0} & \usebox{1}
\end{tabular}
}
\caption{\small{Scatter plot (left) and contour map (right) of a random sample of $(g_1,g_2,g_3)$ for five-thousand $Q \in \mathcal{S}_{(12,3)}$ on the Heart Failure Clinical Records data set ($\hat{p} = 0.75$).} The color in both plots is determined by $g_3$, where darker color corresponds to low $g_3$ (high cost). Note the trade-off between conclusive accuracy and cost: $g_3$ is relatively low in the bottom-right corner of the contour plot and relatively high in the top-left corner.}
\label{fig:landscape_backup}
\end{center}
\end{figure}

    Objectives are designed to incorporate important performance aspects of a real-time detection system. Leveraging the mathematical flexibility of EAs, we can incorporate pathological functions (e.g., indicators) and define several objectives separately. All objectives are to be \textit{maximized} in the range $[0,1]$. See Figure \ref{fig:landscape_backup} for an example visual depiction of the fitness landscape determined by these objectives.
    In the following subsections, $m$ denotes the index of the final stage at which input $\mathbf{x}$ is processed.

\begin{lrbox}{0}
\begin{tikzpicture}[]
\begin{axis}[view/h=-20,
	colormap/viridis, 
	grid style={line width=.15pt, draw=gray!30},
	major grid style={line width=.1pt,draw=gray!50},
	ylabel=$g_1$, 
	xlabel=$g_2$,
	zlabel=$g_3$,
	view={8}{28},
	legend pos=north west,
	grid=both,
    minor x tick num={4},
    minor y tick num={4},
    minor z tick num={4},
    enlargelimits={abs=0},
    % ticklabel style={font=\tiny,fill=white},
    axis line style={latex-latex}
]
\addplot3[scatter, only marks, mark size=1pt, each nth point={1}] table[] {heart1.dat};  
% \addplot3[mesh/rows=70,mesh/num points=5000,contour gnuplot={
%                     number=24,
%                     % cdata should not be affected by z filter:
%                     output point meta=rawz,
%                     labels=false,
%                 }] table [] {heart1.dat};
\end{axis}
\end{tikzpicture}
\end{lrbox}

\begin{lrbox}{1}
\input{heart_contour_30.tex}
\end{lrbox}

    \subsubsection{Conclusiveness} \label{sec:conc}
    This objective attempts to quantify an abstract notion of \textit{utility} representing the insight provided by the classification system under a minimum confidence constraint. Specifically, conclusiveness is measured as the fraction of predictions that meet a targeted confidence threshold, $\hat{p}$.
    \begin{equation}\label{conc}
    g_1(Q) = \frac{1}{N}\sum_{\mathbf{x} \in \mathcal{X}_{test}} \mathbbm{1}[\bar{p_m} \geq \hat{p}]
    \end{equation}
    If conclusiveness is not considered as an objective, then a risk-averse monitor accepting only confident predictions may obtain little information from the system. Such a monitor wishes to be correct at a rate at least $\hat{p}$ but also desires that the classification system yield sufficient information and not terminally reject an excessive fraction of predictions.
    
    Such confidence-constrained classification is common in practice \citep{cancerconf,patel}. Despite being an important aspect of performance that has presence in the literature outside of budgeted learning \citep{cortes,geifman2017}, conclusiveness (coverage) is not accounted for explicitly in current budgeted learning methods, to the best of our knowledge. Note that reject decisions can be defined in a number of ways \citep{yaniv10,geifman2017}. \texttt{EMSCO} utilizes a rejection protocol based on confidence thresholds because it is computationally efficient and frequently used in practice. It is also intuitive; when a well-calibrated classification algorithm (e.g., logistic regression) is used at each stage, a reject protocol based on confidence thresholds may effectively ensure accuracy above $\hat{p}$ (Figure \ref{fig:threshold}). Note, however, poorly calibrated classification algorithms (i.e., those with large disparities between confidence scores and actual class probabilities) may obscure the interpretation of this rejection protocol \citep{kuhn2013}.
    \subsubsection{Conclusive Accuracy}
     Conclusive accuracy is the proportion of conclusive, correctly classified inputs with respect to the total number of conclusive predictions. Let $y$ denote the true label for $\mathbf{x} \in \mathcal{X}_{test}$, and $\bar{\mathscr{C}_m}(Q_m,\mathbf{x})$ return the label corresponding to the maximum class probability after evaluation in stage $m$. Conclusive accuracy is defined as:
    \begin{equation} \label{acc}
    g_2(Q) = \frac{1}{N^*} \sum_{\mathbf{x} \in \mathcal{X}_{test}} \mathbbm{1}[\bar{\mathscr{C}_m}(Q_m,\mathbf{x}) = y \And \bar{p_m} \geq \hat{p}],
    \end{equation}
    where $N^*$ denotes the number of conclusive predictions. By default, $g_2$ applies the same penalty for all misclassification types. However, it can be easily modified to penalize different cases (e.g., false negative, false positive) with a piecewise function.

    \subsubsection{Cost} \label{sec:cost}
    Because feature acquisition typically comprises the bulk of test-time processing expense \citep{kusner}, raw cost $g_3^*(\cdot)$ is measured as the \textit{average sum of acquisition costs per test input}:
    $$g_3^*(Q) = \frac{1}{N}\sum_{\mathbf{x} \in \mathcal{X}_{test}} \left(\sum_{j = 1}^{j=m} \mathcal{C}_{Q_j}\right)$$
    where $\mathcal{C}_{Q_j}$ denotes the sum of features' costs in stage $j$. Note that the cost of inconclusive records is considered since the system requires resources to arrive at an inconclusive result. \\\\
    To maintain consistency as a maximization problem and a [0,1] scale for objectives, the minimum raw cost of all solutions in the current generation is determined and used for normalization. Then, for $Q \in G_h$, inverse cost $g_3(Q)$, is measured as:
    \begin{equation}\label{cost} 
        g_3(Q) = \frac{\min_{U \in G_h} g^*_3(U)}{g^*_3(Q)}
    \end{equation}

    \subsection{Problem Definition} \label{problemdef}
    \texttt{EMSCO} seeks an ordered feature set partition, $Q \in \mathcal{S}_{(n,k)}$, to maximize objectives $g_1,g_2,g_3$ through the perspective of \textit{Pareto efficiency}. Consider two chromosomes $Q_A, Q_B$. If every $g_i(Q_A) \geq g_i(Q_B)$, and there is at least one objective such that $g_i(Q_A) > g_i(Q_B)$, then $Q_A$ is said to \textit{dominate} $Q_B$, and this relationship is denoted as $Q_A \succ Q_B$. A solution that is not dominated by any solution in an arbitrary population $G$ is said to be \textit{non-dominated} and solves its respective multi-objective optimization problem \citep{deb01}. Note, $\mathscr{N}(G)$ returns the set of all non-dominated solutions in $G$, and this set of non-dominated solutions is referred to as the \textit{Pareto frontier} of population $G$.
    
    \texttt{EMSCO} addresses the following multi-objective optimization by seeking solutions that are non-dominated in $\mathcal{S}_{(n,k)}$:
    \begin{equation} \label{problem}
        \argmax_{Q \in \mathcal{S}_{(n,k)}} \left(g_1(Q),g_2(Q),g_3(Q)\right).
    \end{equation}
    In this manuscript, solutions to this problem are referred to as \textit{globally Pareto optimal} or \textit{globally non-dominated}.

    \section{Evolving Multi-Stage Classifier Configurations} \label{sec:emsco}
    We propose a problem-specific evolutionary algorithm, \textit{Evolutionary Multi-Stage Classifier Optimizer} (\texttt{EMSCO}), that seeks to solve the optimization problem in (\ref{problem}). \texttt{EMSCO} evolves improved populations of configurations over a series of generations (iterations of the algorithm). Key aspects of \texttt{EMSCO} include the chromosomal representation of solutions and the evolutionary operators employed to create new populations. These are detailed in the sections below.

    \subsection{Chromosome Representation of Ordered Feature Set Partitions}\label{representation}

    Ordered feature set partitions can be conveniently represented as lists. Let $Q$ be a solution in $\mathcal{S}_{(n,k)}$. Then the chromosomal representation of $Q$ is: $$[Q] \in \{0,1,\ldots,k-1\}^n,$$ where $[Q]_j$ corresponds to the $j$th feature's stage assignment. The elements in $[Q]$ are \textit{zero-indexed}; so $[Q]_j = s$ assigns feature $j$ to stage $s+1$.  For example, given the ordered set partition $Q = \left \{ \{\mathcal{F}_1, \mathcal{F}_2\}, \{\mathcal{F}_3,\mathcal{F}_4\} \right \}$, the corresponding list representation would be $[Q] = \{0,0,1,1\}$. 

     There is not an exact correspondence between list representations and ordered feature set partitions. For example, $\{0,0,0,2\} \in \{0,1,2\}^4$, but since stage two is empty, this list does not correspond to an ordered feature set partition and is not a feasible solution. Asymptotically, the disparity between these sets becomes less problematic (Theorem \ref{thm:solutionspace}), but a remedy is still needed for low dimension data sets. 
    
    After a new solution is generated, it is checked for gaps with the function, $gaps([C])$, that returns true if any $j=0...\Norm{C}-1$ is missing. If any such gaps exist, the solution is modified with $compress([C])$ to redefine stage assignments in the following manner: While there is a missing $j^* \in \{0,\ldots,\Norm{C}-1\}$, iterate through $[C]$ and replace $j^* + 1$ with $j^*$. For instance, in the above example, $\{0,0,0,2\}$ would be compressed to $\{0,0,0,1\}$.

    \subsection{Evolutionary Processes}\label{sec:operators}
    \texttt{EMSCO}'s evolutionary operators are designed to leverage problem-specific knowledge while maintaining breadth of search to avoid premature convergence. These operators are applied in sequence to form a new population per generation. 

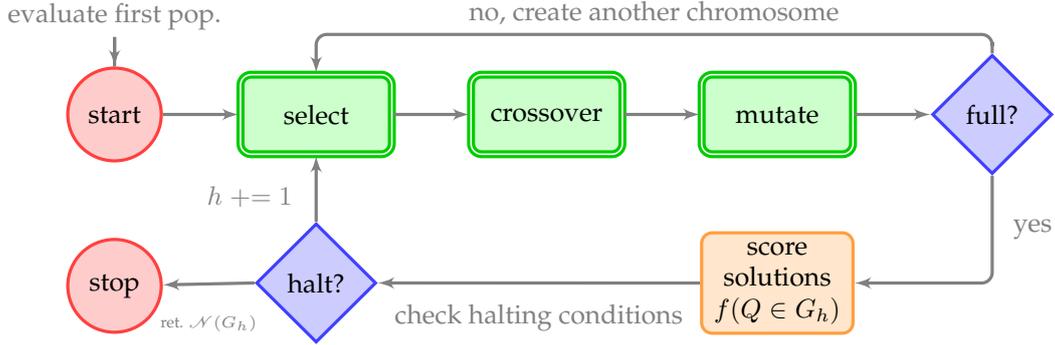
\begin{figure}
\begin{center}
% \begin{figure*}
% \begin{center}
% \resizebox{5.25in}{!}{
% \fbox{
% \input{camProc}  
\begin{tikzpicture}% [node distance = 3.25cm, 
[align = center]
% \draw[black,very thick, help lines] (0, 0) grid (18, 8);

\node[draw = red!75, fill = red!20, circle, very thick, minimum height = 3.5em] at (0, 4) (start) {start};
\node[draw = green!80!black, fill = green!20, rectangle, very thick, double, text width = 5em, text centered, rounded corners, minimum height = 3em, right = 1cm of start] (select) {select};
\node[draw = green!80!black, fill = green!20, rectangle, very thick, double, text width = 5em, text centered, rounded corners, minimum height = 3em, right = 1cm of select] (crossover) {crossover};
\node[draw = green!80!black, fill = green!20, rectangle, very thick, double, text width = 5em, text centered, rounded corners, minimum height = 3em, right = 1cm of crossover] (mutate) {mutate};
\node[draw = orange!75, fill = orange!20, rectangle, very thick, text width = 5em, text centered, rounded corners, minimum height = 3em, below = of mutate] (evaluate) {score solutions \\$f(Q \in G_h)$};
\node[draw = red!75, fill = red!20, circle, very thick, minimum height = 3.5em, below = of start] (stop) {stop};

\node[draw = blue!75, fill = blue!20, diamond, very thick, text width = 3.75em, text badly centered, inner sep = 0pt, right = of mutate] (full) {full?};
\node[draw = blue!75, fill = blue!20, diamond, very thick, text width = 3.75em, text badly centered, inner sep = 0pt, yshift = 0.125cm, below = of select] (done) {halt?};

% nodes to help the "fit" rectangles 
\node[draw = none, below = of full] (belowfull) {};

% \node[draw = black, dashdotted, inner sep = 10pt, fit = (chromosomeNW) (chromosomeSE)](chromosome) {};

\path[draw, very thick, color = black!50, -latex'] (start) -- (select);
\path[draw, very thick, color = black!50, -latex'] (select) -- (crossover);  
\path[draw, very thick, color = black!50, -latex'] (crossover) -- (mutate);
\path[draw, very thick, color = black!50, -latex'] (mutate) -- (full);
% \path[draw, very thick, color = black!50, -latex'] (sfull) -- node[label = west:5 \shortstack{\\ yes, new \\ chromosome \\ generated}]{} (evaluate);

\path[draw, very thick, rounded corners, color = black!50, -latex'] (full.south) |- node[near start, label = east: yes]{} (evaluate.east);

\path[draw, very thick, rounded corners, color = black!50, -latex'] (full.north) -| +(0, 0.25) -| node[above, near start]{no, create another chromosome} (select.north);

\path[draw, very thick, rounded corners, color = black!50, -latex']
(start.north) -| +(0, 0.25) -| node[above, near start, yshift = 0.1cm]{evaluate first pop.} (start.north);

\path[draw, very thick, rounded corners, color = black!50, -latex'] (evaluate.west) -- node[midway, label = south: {\shortstack{check halting conditions % \\ create new population}}]{} (done.east);
}}]{} (done.east);

% \path[draw, very thick, rounded corners, color = black!50, -latex'] (done.west) -- ++(-1, 0) node[left, xshift = -0.1cm]{yes};
\path[draw, very thick, rounded corners, color = black!50, -latex'] (done.west) --  node[midway, below, label = south:{\tiny{ret. $\mathscr{N}(G_h)$}}]{} (stop);

\path[draw, very thick, rounded corners, color = black!50, -latex'] (done.north) -- node[left, xshift = 0.1cm, yshift = -0.1cm, label = west: {\shortstack{$h \pluseq 1$}}]{} (select.south);

\end{tikzpicture}
% }
% \caption{Flow chart of the tasks used for identifying multi-stage classifier configurations. Double-lined blocks (also in green) contain traditional EA processes, while blocks in orange are part of the configuration assessment. }
% \label{fig:eaProc}
% \end{center}
% \end{figure*}
\caption{\small{Flow chart of the tasks used for identifying MSC configurations. Double-lined blocks (in green) contain traditional GA processes, while the orange block performs MSC configuration assessment (ranking and fitness).}}
\label{fig:ea}
\end{center}
\end{figure}

\subsubsection{Selection} \label{sec:selection}

Selection is the process in which individual solutions (chromosomes) are chosen from a population for the recombination stage. \texttt{EMSCO} utilizes roulette wheel selection \citep{eiben}, which biases the selection with respect to the fitness metric (described in Section \ref{sec:fitness}). 
As a result, more fit chromosomes are more likely to be chosen.

\subsubsection{Elitism}
The elitism protocol applied in this paper is designed to preserve all \textit{unique} non-dominated solutions to the subsequent generation. Let $E_0^*$ denote the set of unique non-dominated chromosomes within generation $G_h$, and let $G_h^*$ denote the set of unique solutions in generation $G_h$. 
For elitism parameter $b$ in the open interval $(0,1)$, the top $M = \max\left(\lceil b\Norm{G_h^*}\rceil, \Norm{E_0^*}\right)$ solutions in each generation are sent to the subsequent generation without modification. %As a halting condition, if $M \geq \Norm{G}$, \texttt{EMSCO} returns the entire elite population and stops execution. This helps ensure the algorithm does not discard non-dominated solutions (See Section \ref{sec:algorithm}).

\subsubsection{Mutation}\label{sec:mutation}

In general, mutation operators are used to maintain genetic diversity from one generation's population of chromosomes to the next \citep{eiben}. \texttt{EMSCO} uses mutation to implement this behavior while also addressing certain domain-specific concerns. While traversing a newly created chromosome, mutation occurs with probability $\hat{m}$ at each index.

If performance is comparable between solutions $Q_A,Q_B \in G_h$, the solution with lower stage count is preferred as it will likely impose fewer classifier evaluation costs than the alternative, and these costs can prove consequential in high-dimensional settings or when expensive classification algorithms are employed at each stage \citep{block}. Moreover, in marginal cases, each additional stage in an MSC presents another opportunity to trigger a wrong but conclusive prediction. An incorrect prediction only needs to be declared conclusive once in order to be recorded definitively (Section \ref{sec:setup}), and the probability of this event may increase as more trials (stages) are introduced. 

With these aspects in mind, we prefer that solutions with high stage count not be considered until lower stage counts have proven to be insufficient in comparison. To this end, a discrete, monotonically decreasing probability distribution is desired to mutate chromosomes' stage assignments. The beta-binomial distribution \citep{casella01} with $\alpha =1 \And \beta > \alpha$, denoted $BB(\alpha=1,\beta)$, satisfies this criteria and possesses several convenient properties for adjusting bias of the mutation operator towards low stage assignments. Let $0 \leq j \leq \Norm{Q}$ be a potential stage assignment for the $i$\textsuperscript{th} feature in chromosome $Q$. The probability mass function is given by:
\begin{equation} \label{eqn:mutation3}
    \mathbb{P}\left [ ~[Q]_i \gets j~ \right ] = \binom{\Norm{\textup{\small{Q}}}}{j} \frac{B\left (j + 1, \Norm{\textup{\small{Q}}}-j+\beta\right )}{B\left (1, \beta\right)}
\end{equation}
where $B(\cdot)$ denotes the beta function. For a chromosome with stage count $\Norm{Q}$, this distribution has an expected value ($\frac{\Norm{Q}}{\beta + 1}$) that is decreasing with respect to $\beta$. This parameter can therefore be used to adjust bias, where greater $\beta$ corresponds to greater preference for low stage assignments.
    
Despite the bias described above, it is important that the mutation operator allows creation of new stages when warranted---in certain scenarios (e.g., high $n$), a low stage count may yield high feature acquisition cost at each stage. The mutation operator creates new stages in $[Q]$ by assigning stage $\Norm{Q}$ to one of the elements \footnote{Recall, the list representation is zero-indexed at each element, so a stage assignment of $\Norm{Q}$ sends the feature to new stage $\Norm{Q} + 1$}, and the probability of adding a stage in this manner can be controlled by increasing/decreasing $\beta$. See Algorithm \ref{alg:mutation} for a pseudo-code representation of the mutation procedure.

    \begin{algorithm}
    \SetAlgoLined
    $Q$ is a new chromosome\\
    \tcp{$0 \leq i < n$}
    \For{\textup{each} $i$\textup{\textsuperscript{th}} \textup{feature}} {

    \If{$\left(rand([0,1]) \leq \hat{m} \And count([Q]_i, [Q]) > 1\right)$} {
    $[Q]_i \sim BB(\min \left(\Norm{\textup{\small{Q}}},k-1\right),1, \beta)$; 
    }
    }
    \caption{Mutation}
    \label{alg:mutation}
    \end{algorithm}
    The condition $count([Q]_i, [Q]) > 1$ checks if there is more than one occurrence of $[Q]_i$ in $[Q]$ and is considered to ensure that mutation will not result in an empty stage. The draw from $\min \left(\Norm{\textup{\small{Q}}},k-1\right)$ elements ensures that the maximum number of stages $k$ is not exceeded.

    \subsubsection{Recombination}

    \texttt{EMSCO}'s recombination/crossover operator takes into account the differing contexts of stage assignments within the parent solutions. To implement crossover, two parent chromosomes $P_A, P_B \in G_h$ are selected according to the protocol described in Section \ref{sec:selection}. With probability $\hat{r}$, these solutions are then combined to produce a child solution, $C \in G_{h+1}$. With probability $1 - \hat{r}$, recombination/crossover returns one of $P_A,P_B$ via a uniform random selection (coin flip).
    
     The first step in recombination of $P_A$ and $P_B$ chooses a stage count for the child solution. Three options exist for $\Norm{C}$: $\Norm{P_A}, \Norm{P_B}, \left \lfloor \frac{\Norm{P_A}+\Norm{P_B}}{2}\right\rfloor$, where the third option is incorporated to facilitate consideration of solutions with intermediate stage counts and introduce a notion of blending parent chromosomes with respect to stage count. After a stage count is decided, each of the child's features must be assigned a stage. Each $i$th assignment is initially chosen to be either $[P_A]_i$ or $[P_B]_i$. Let $R \in \{P_A, P_B\}$ denote the parent corresponding to this choice. In an \textit{attempt} to maintain the context of the parent's stage assignment for the $i$th feature, the initial selection $[R]_i$ is normalized according to $\Norm{R}$ to acquire the relative order of the stage assignment within $R$. This ratio is then multiplied by $\Norm{C}$, rounded, and decremented to obtain a zero-indexed, approximate analog of $[R]_i$ in $[C]$.
     Alternatively, if raw values were assigned ($[C]_i \gets [R]_i$), then the chosen stage assignment may lose context in the child solution's structure. Consider, for example, the role of the second stage in a five-stage system versus the role of the second stage in a two stage system. In the latter case, the second stage provides the final evaluation of inputs; In the former case, the second stage is an early evaluation. The recombination process is detailed in Algorithm \ref{alg:recombination}.

\begin{figure}
\begin{center}
% \begin{figure}
% \begin{center}
% \resizebox{5.25in}{!}
{
% \fbox{
% \input{camProc}  
\begin{tikzpicture}% [node distance = 3.25cm, 
[align = center]
% \draw[black, very thick, help lines] (0, 0) grid (6, 3);

\node[draw, blue, rectangle, minimum size = 0.5cm, fill = blue!20, label=north:{$\mathcal{F}_{1}$}, label=west:{$[P_A]=$}] at (0, 3) (a1) {};
\foreach \x [remember=\x as \lastx (initially 1)] in {2,...,5}{
    \node [draw, blue, rectangle, fill = blue!20, right = 0cm of a\lastx.east, 
           minimum size = 0.5cm, label=north:{$\mathcal{F}_{\x}$}] (a\x) {};
}

\node[]at (a1.center) {$0$};
\node[]at (a2.center) {$2$};
\node[]at (a3.center) {$1$};
\node[]at (a4.center) {$2$};
\node[]at (a5.center) {$1$};
% \node[draw = none, rectangle, minimum size = 0.5cm, left = 0.0cm of a1] {$L_{1} =$};

\node[right = 3cm of a5, draw, blue, rectangle, minimum size = 0.5cm, fill = blue!20, label=north:{$\mathcal{F}_{1}$}]  (na1) {};
\foreach \x [remember=\x as \lastx (initially 1)] in {2,...,5}{
    \node [draw, blue, rectangle, fill = blue!20, right = 0cm of na\lastx.east, 
           minimum size = 0.5cm, label=north:{$\mathcal{F}_{\x}$}] (na\x) {};
}

\node[]at (na1.center) {$0$};
\node[]at (na2.center) {$1$};
\node[]at (na3.center) {$0$};
\node[]at (na4.center) {$1$};
\node[]at (na5.center) {$0$};

\node[below = 2cm of a1, draw, red, rectangle, minimum size = 0.5cm, fill = red!20, label=south:{$\mathcal{F}_{1}$}, label=west:{$[P_B]=$}] (b1) {};
\foreach \x [remember=\x as \lastx (initially 1)] in {2,...,5}{
    \node [draw, red, rectangle, fill = red!20, right = 0cm of b\lastx.east, 
           minimum size = 0.5cm, label=south:{$\mathcal{F}_{\x}$}] (b\x) {};
}

\node[]at (b1.center) {$1$};
\node[]at (b2.center) {$0$};
\node[]at (b3.center) {$0$};
\node[]at (b4.center) {$1$};
\node[]at (b5.center) {$0$};
% \node[draw = none, rectangle, minimum size = 0.5cm, above = 0.0cm of b1] {$L_{2} =$};

\node[right = 3cm of b5, draw, red, rectangle, minimum size = 0.5cm, fill = red!20, label=south:{$\mathcal{F}_{1}$}]  (nb1) {};
\foreach \x [remember=\x as \lastx (initially 1)] in {2,...,5}{
    \node [draw, red, rectangle, fill = red!20, right = 0cm of nb\lastx.east, 
           minimum size = 0.5cm, label=south:{$\mathcal{F}_{\x}$}] (nb\x) {};
}

\node[]at (nb1.center) {$1$};
\node[]at (nb2.center) {$0$};
\node[]at (nb3.center) {$0$};
\node[]at (nb4.center) {$1$};
\node[]at (nb5.center) {$0$};

\coordinate (a5b1) at ($(a5)!0.5!(b1)$);
\coordinate (na5nb1) at ($(na5)!0.5!(nb1)$);

\node[draw = green!80!black, rounded corners, thick, rectangle, minimum size = 0.5cm, fill = green!20] at ($(a5)!0.5!(na1)$) (norma) {\shortstack{normalize \\ stages}};

\path[draw, black!50, very thick, rounded corners, latex'-] (norma.north) -- +(0, 0.5) node[yshift = 0.25cm]{\textcolor{black}{$\Norm{C}$}};

\node[draw = green!80!black, rounded corners, thick, rectangle, minimum size = 0.5cm, fill = green!20] at ($(b5)!0.5!(nb1)$) (normb) {\shortstack{normalize \\ stages}};

\path[draw, black!50, very thick, rounded corners, latex'-] (normb.north) -- +(0, 0.5) node[yshift = 0.25cm]{\textcolor{black}{$\Norm{C}$}};

% \coordinate (na5nb5) at ($(na5)!0.5!(nb5)$);

\node[right = 2cm of na5, draw = blue, rectangle, minimum size = 0.5cm, fill = blue!20, label=north:{$\mathcal{F}_{1}$}, label=west:{$[C]=$}]  (c1) {0};
\node[right = 0cm of c1.east, draw = red, rectangle, minimum size = 0.5cm, fill = red!20, label=north:{$\mathcal{F}_{2}$}]  (c2) {0};
\node[right = 0cm of c2.east, draw= blue, rectangle, minimum size = 0.5cm, fill = blue!20, label=north:{$\mathcal{F}_{3}$}]  (c3) {0};
\node[right = 0cm of c3.east, draw = red, rectangle, minimum size = 0.5cm, fill = red!20, label=north:{$\mathcal{F}_{4}$}]  (c4) {1};
\node[right = 0cm of c4.east, draw = red, rectangle, minimum size = 0.5cm, fill = red!20, label=north:{$\mathcal{F}_{5}$}]  (c5) {0};

\path[draw, very thick, rounded corners, color = blue!50, -latex'] (a5.east) -- (norma.west);
\path[draw, very thick, rounded corners, color = blue!50, latex'-] (na1.west) -- (norma.east);
\path[draw, very thick, rounded corners, color = red!50, -latex'] (b5.east) -- (normb.west);
\path[draw, very thick, rounded corners, color = red!50, latex'-] (nb1.west) -- (normb.east);

\path[draw, very thick, rounded corners, color = blue!50, -latex'] (na1.south) -- ++(0.0, -0.5) -| (c1.south);
\path[draw, very thick, rounded corners, color = blue!50, -latex'] (na3.south) -- ++(0.0, -0.75) -| (c3.south);
\path[draw, very thick, rounded corners, color = red!50, -latex'] (nb2.north) -- ++(0.0, 0.25) -| (c2.south);
\path[draw, very thick, rounded corners, color = red!50, -latex'] (nb4.north) -- ++(0.0, 0.5) -| (c4.south);
\path[draw, very thick, rounded corners, color = red!50, -latex'] (nb5.north) -- ++(0.0, 0.75) -| (c5.south);

\draw [decorate, decoration={brace, amplitude=6pt}, xshift = 0pt, yshift = 0pt]
($(na5.north east) + (0, 0.75)$) -- ($(c1.north west) + (0, 0.75)$) node [black, midway, yshift = 0.65cm]
{\footnotesize \shortstack{random selection of $\mathcal{F}_{i}$ \\ from $[P_{A}]$ and $[P_{B}]$} };

\end{tikzpicture}

% }
}
% \caption{Example stage normalization and recombination, where $[P_{A}]$ and $[P_{B}]$ are parents and $[C]$ is the child. }
% Double-lined blocks (also in green) contain traditional EA processes, while blocks in orange are part of the configuration assessment.}
% \label{fig:recombineH}
% \end{center}
% \end{figure}
\caption{\small{Example stage normalization and recombination, where $[P_{A}]$ and $[P_{B}]$ are parents and $[C]$ is the child. In this example, the child's stage count is selected as $\Norm{P_B}$.}}
\label{fig:recombine}
\end{center}
\end{figure}
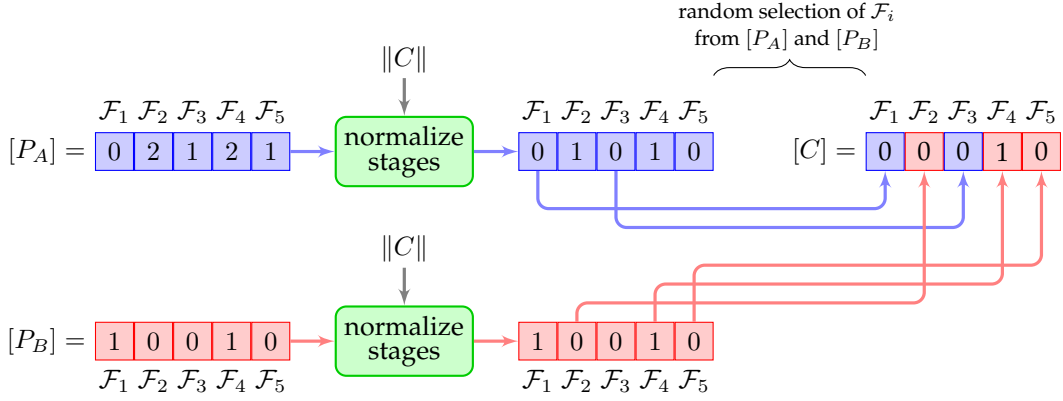

    \begin{algorithm}
    \SetAlgoLined
    $[P_A]$, $[P_B]$ are parents selected for recombination in $G_h$\\
    Initiate child $[C] \gets [0,0,\ldots,0]$\\
    $\Norm{C} \gets rand\left \{ \left \lfloor \frac{\Norm{P_A} + \Norm{P_B}}{2}\right \rfloor, \Norm{P_A}, \Norm{P_B} \right \}$;\\
    \tcp{$0 \leq i < n$}
    \For{\textup{each} $i$\textup{\textsuperscript{th}} \textup{feature}} {
     $R = rand \left \{P_A, P_B \right \}$;\\
     Assign stage to the $i$\textsuperscript{th} feature of the child\\
     $[C]_i \gets_{round} \left (\frac{[R]_i + 1}{\Norm{R}}  \right) \Norm{C}$;\\ 
     $[C]_i \gets [C]_i - 1$;\\
     $[C]_i \gets \max \left([C]_i,0\right)$;
     
    }
    \Return{$[C]$}
    \caption{Recombination}
    \label{alg:recombination}
    \end{algorithm}

    \subsection{Evaluating Chromosomes}\label{sec:fitness}
    For a solution $Q \in G_h$, fitness is computed using the $\ell_2$ norm of $\left\{g_1(Q),g_2(Q),g_3(Q)\right\}$, denoted by $\mathscr{E}(Q)$, and an exponential term accounting for the \textit{Pareto rank} of $Q$ with respect to the current generation. In this paper, Pareto rank, or \textit{non-domination level} \citep{996017}, is an indication of the solution's performance from the perspective of Pareto efficiency. This mixed formulation of fitness respects Pareto efficiency and raw aggregated performance simultaneously. 
    \subsubsection{Pareto Rank of Chromosomes}
   To compute $rank(Q)$ in generation $G_h$, we first determine the non-dominated solutions in $G_h$.  This set of solutions $E_0$ is then removed from $G_h$. The non-dominated solutions in set $G_h \setminus E_0$ are then stored in the set $E_1$. This process of removing non-dominated solutions continues so that $E_t$ contains the non-dominated solutions in the population $G_h \setminus E_0 \setminus E_1 \setminus \ldots \setminus E_{t-1}$. Once all solutions have been assigned to some $E_t$, the ranking procedure is finished. 
    \
    Let $E_{t^*}$ denote the final non-dominated set removed from $G_h$, and let $Q \in E_t$. Then 
    \begin{equation}\label{rank}
    rank(Q) = t^* - t
    \end{equation}
    Let $R,Q \in G_h$ be chromosomes in the $h$\textsuperscript{th} generation. $rank(R) > rank(Q)$ does not necessarily imply $R \succ Q$---it only implies that there exists some chromosome $S \in G_h$ such that $S \succ Q$ and  $S \not \succ R$. Note that our assignment of non-domination level is flipped---typically, the first non-dominated set is assigned zero rank \citep{996017}. However, we define $rank(\cdot)$ so that the initial (best) set of non-dominated solutions is assigned the greatest value. The motivation for this may become clearer in the next subsection, where we define fitness.
    \subsubsection{Fitness Function}
    Fitness of chromosome $Q$ is computed as a product involving $rank(Q)$ and the $\ell_2$ norm, $\mathscr{E}(Q)$. Let $l_\mathscr{E}$ and $u_\mathscr{E}$ denote the minimum and maximum values of $\mathscr{E}(\cdot)$ in generation $G_h$, respectively. For some $\epsilon > 0$, we set $$\gamma=\frac{u_\mathscr{E}}{l_{\mathscr{E}}} + \epsilon.$$ Fitness is defined as:
     \begin{equation}\label{formula:fitness}
         f(Q) = \gamma^{rank(Q)}\sqrt{g_1(\small{Q})^{2} + g_2(\small{Q})^{2} + g_3(\small{Q})^{2}},
     \end{equation}
     By default, \texttt{EMSCO} uses $\epsilon=0.01$, but this value can be modified to any positive real value. Rather than use a linear weighted-sum, the $\ell_2$ norm is incorporated for its stability and to introduce a visual interpretation of aggregated performance as distance from the origin. This formulation of fitness contributes to several useful properties of \texttt{EMSCO} discussed in the next section.
    Note, the global \textit{scalarized} optimization problem posed by this fitness function is:
    \begin{equation} \label{scalarization}
        \argmax_{Q \in \mathcal{S}_{(n,k)}} \large{f}(g_1(Q),g_2(Q),g_3(Q))
    \end{equation}    

    \subsection{Algorithmic Details and Guarantees} \label{sec:algorithm}
    \texttt{EMSCO} begins by generating a random initial population of $\Norm{G}$ chromosomes by calling the function $init()$, which creates solutions by applying the mutation operator to the default one-stage solution ($[0,0,\ldots,0]$). After the first population has been generated, the main loop begins. Two halting conditions are checked before each iteration by calling the function, $converged()$, which returns true if either of the following conditions are satisfied---(i) the maximum number of iterations ($max\_iter$) have been executed or (ii) the highest-scoring chromosome has remained constant for the past $g$ generations. If neither of these halting conditions are true, the loop proceeds, and the next population of solutions is produced using elitism, selection, recombination, and mutation as described in Section \ref{sec:operators}. When the loop ends, the list of all non-dominated solutions in the final generation is returned, sorted by fitness.
    \begin{algorithm} \label{alg:emsco}
    \SetAlgoLined
    $G_0 \gets init()$;\\ 
    $h \gets 0$\\
    \While{not $converged()$} {
     $rank(G_h)$;
     $sort\left(G_h,key=f(\cdot)\right)$;\\
     $elite\_size \gets \max\left(\lceil b\Norm{G_h^*}\rceil, \Norm{E_0^*}\right)$;\\
     $elite\_pop \gets G_h^*\left[0:elite\_size\right]$;\\
     %\If{$elite\_size \geq pop\_size$} {\Return{$elite\_pop$;}}
     $G_{h+1} \gets_{append} elite\_pop$;\\
     \While{$\Norm{G_{h+1}} < \Norm{G}$} 
     {
     $P_A,P_B \gets select()$;\\
     $[C] \gets recombine([P_A],[P_B]);$\\
     $[C] \gets mutate([C])$;\\
     $G_h \gets_{append} C$;\\
     }
     $h \gets h+1$;\\}
     \Return{$\mathscr{N}(G_h)$};\\
    
    \caption{\texttt{EMSCO}}
    \end{algorithm}
    Several important properties of \texttt{EMSCO} follow as consequences of the design aspects described heretofore.\\~\\
    
    \begin{property}\label{thm:pres}
    \texttt{EMSCO} preserves globally non-dominated solutions and returns them at the terminal generation.
    \end{property}
    \begin{proof}
    Suppose for contradiction that some globally non-dominated chromosome $D$ was encountered in $G_h$ but not returned at terminal generation $G_{term}$. That is, $D \in G_h$ and $D \not \in \mathscr{N}(G_{term})$.
    
    As a consequence of \texttt{EMSCO}'s elitism protocol, $D$ must have been dominated in some population $G_h, G_{h+1},\ldots,G_{term}$ to have been discarded. This contradicts our initial supposition, since a globally non-dominated solution in $\mathcal{S}_{(n,k)}$ will likewise be non-dominated in any \textit{subset} of $\mathcal{S}_{(n,k)}$.
    \end{proof}
    \begin{property}
     Let $Q \in G_h$. (i) Probability of selection is strictly increasing with respect to $rank(Q)$, and, (ii) within a particular rank, probability of selection is strictly increasing with respect to $\mathscr{E}(Q)$.
    \end{property}
    \begin{proof}
     ~\\(i) Let $R_A,R_B$ be chromosomes in generation $G_h$ such that $$r_B = rank(R_B) > r_A = rank(R_A).$$ Because $\frac{u_\mathscr{E}}{l_{\mathscr{E}}}$ maximizes the ratio between any two $\mathscr{E}(R_A)$ and $\mathscr{E}(R_B)$, and we have assumed $r_B > r_A$, we have that $$\left(\frac{u_\mathscr{E}}{l_{\mathscr{E}}} + \epsilon\right)^{r_B - r_A} > \frac{\mathscr{E}(R_A)}{\mathscr{E}(R_B)},$$ since $\gamma^r$ is strictly increasing with respect to $\gamma$ and $r$. The above inequality is equivalent to $$\left(\frac{u_\mathscr{E}}{l_{\mathscr{E}}} + \epsilon \right)^{r_A}\mathscr{E}(R_A) < \left(\frac{u_\mathscr{E}}{l_{\mathscr{E}}} + \epsilon \right)^{r_B}\mathscr{E}(R_B)$$ which, by definition, implies $$f(R_A) < f(R_B).$$
     The result follows from the use of roulette wheel selection. \\(ii).  If $r_A = r_B$, fitness comparisons between $R_A,R_B$ are determined exclusively by $\mathscr{E}(\cdot)$. 
    \end{proof}

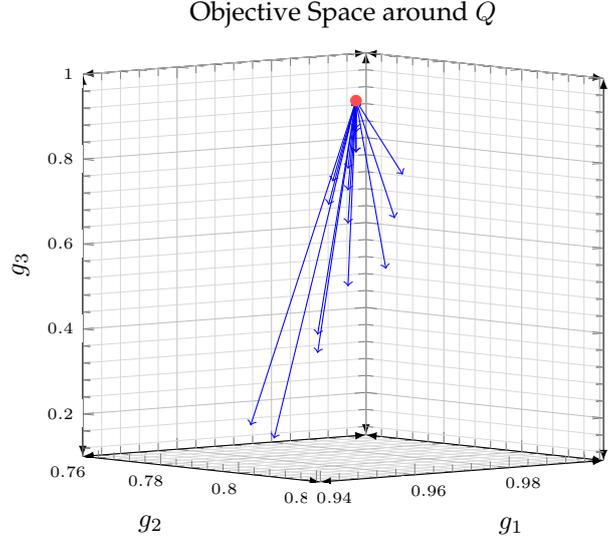
\begin{figure}
 \begin{center}
\begin{tikzpicture}

\begin{axis}[grid style={line width=.15pt, draw=gray!30},xmin=.76,xmax=.821,ymax=1,ymin=.94,zmin=.1,zmax=1,major grid style={line width=.1pt,draw=gray!50},title={Objective Space around $Q$},ylabel=$g_1$,xlabel=$g_2$,zlabel=$g_3$,view={50}{5},legend pos=north west,grid=both,
    minor x tick num={4},
    minor y tick num={4},
    minor z tick num={4},
    enlargelimits={abs=0},
    ticklabel style={font=\tiny,fill=white},
    axis line style={latex-latex}]
 
%plot in all axes

%lot accuracy vs. cost (x= g_2 vs. z = g_3)
\addplot3[mark=*, only marks, color=red!70] table [x = col1, y = col2,  z = col3] 
{
col1 col2 col3
.772 .988 .909
};

\draw[<-,blue] (0.772,0.988,0.909)--(0.772,0.988,0.909) node[right]{};
\draw[<-,blue] (0.772,0.988,0.787)--(0.772,0.988,0.909) node[right]{};
\draw[<-,blue] (0.772,0.988,0.913)--(0.772,0.988,0.909) node[right]{};
\draw[<-,blue] (0.772,0.988,0.834)--(0.772,0.988,0.909) node[right]{};
\draw[<-,blue] (0.772,0.988,0.913)--(0.772,0.988,0.909) node[right]{};
\draw[<-,blue] (0.772,0.988,0.787)--(0.772,0.988,0.909) node[right]{};
\draw[<-,blue] (0.776,0.983,0.757)--(0.772,0.988,0.909) node[right]{};
\draw[<-,blue] (0.79,0.965,0.352)--(0.772,0.988,0.909) node[right]{};
\draw[<-,blue] (0.766,0.988,0.713)--(0.772,0.988,0.909) node[right]{};
\draw[<-,blue] (0.814,0.931,0.234)--(0.772,0.988,0.909) node[right]{};
\draw[<-,blue] (0.772,0.988,0.909)--(0.772,0.988,0.909) node[right]{};
\draw[<-,blue] (0.772,0.988,0.843)--(0.772,0.988,0.909) node[right]{};
\draw[<-,blue] (0.772,0.988,0.922)--(0.772,0.988,0.909) node[right]{};
\draw[<-,blue] (0.776,0.983,0.628)--(0.772,0.988,0.909) node[right]{};
\draw[<-,blue] (0.772,0.988,0.909)--(0.772,0.988,0.909) node[right]{};
\draw[<-,blue] (0.776,0.983,0.48)--(0.772,0.988,0.909) node[right]{};
\draw[<-,blue] (0.82,0.931,0.209)--(0.772,0.988,0.909) node[right]{};
\draw[<-,blue] (0.771,0.983,0.667)--(0.772,0.988,0.909) node[right]{};
\draw[<-,blue] (0.793,0.977,0.543)--(0.772,0.988,0.909) node[right]{};
\draw[<-,blue] (0.772,0.988,0.909)--(0.772,0.988,0.909) node[right]{};
\draw[<-,blue] (0.772,0.988,0.901)--(0.772,0.988,0.909) node[right]{};
\draw[<-,blue] (0.772,0.988,0.909)--(0.772,0.988,0.909) node[right]{};
\draw[<-,blue] (0.772,0.988,0.901)--(0.772,0.988,0.909) node[right]{};
\draw[<-,blue] (0.784,0.988,0.747)--(0.772,0.988,0.909) node[right]{};
\draw[<-,blue] (0.79,0.965,0.395)--(0.772,0.988,0.909) node[right]{};
\draw[<-,blue] (0.776,0.983,0.706)--(0.772,0.988,0.909) node[right]{};
\draw[<-,blue] (0.788,0.983,0.652)--(0.772,0.988,0.909) node[right]{};
\draw[<-,blue] (0.772,0.988,0.909)--(0.772,0.988,0.909) node[right]{};
%\legend{One-Step Neighborhood,2D $g_2$-$g_3$ trade-off}
\end{axis}

\end{tikzpicture}
\caption{\footnotesize{Example performance landscape of chromosomes within one unit of $[Q]$ (red). Blue vectors map to chromosomes within the $n$-dimensional \textit{von Neumann Neighborhood} \citep{wolfram1} of the red point's  corresponding chromosome. This performance landscape is computed using the Australian Credit Approval data set  (Section \ref{datasets}) with chromosomes in $\mathcal{S}_{(14,4)}$. A single $\pm1$} modification in $[Q]$ may have significant consequences for performance. In this figure, inverse cost is computed globally: $g_3 = \frac{\min_{U \in \mathcal{S}_{(n,k)}} g^*_3(U)}{g^*_3(Q)}$}
\label{fig:neighborhood}
\end{center}

\end{figure}

    \begin{property}
    If \texttt{EMSCO} solves (\ref{scalarization}), then it solves (\ref{problem}).
    \end{property}
    \begin{proof}
     This result follows directly from the construction of fitness\footnote{Note that this result holds for any scalarization that is increasing with respect to all $g_1,g_2,g_3$ \citep{boyd}.}. Any solution that globally maximizes fitness will necessarily be non-dominated.
    \end{proof}
    \texttt{EMSCO} aims to respect both aggregated, scalarized performance ($\mathscr{E}(\cdot)$) and non-domination level/rank. Unlike algorithms such as NSGA-II~\citep{996017}, initial fitness is not determined exclusively by non-domination level, and only the \textit{unique} chromosomes in the first non-dominated front ($E_0^*$) are carried to the next generation automatically. The remaining slots in the subsequent population are filled by executing selection, recombination, and mutation on the entire population in a manner that is biased against but does not exclude low-rank solutions. In our context, low ranked solutions may provide utility under the raw scalarized perspective (high $\mathscr{E}(\cdot)$) and/or be in proximity to high-performing chromosomes in terms of genetic distance. As a cursory demonstration, Figure \ref{fig:neighborhood} depicts the volatility of the solution space near a non-dominated chromosome with respect to genetic distance.
\section{Experiments}
To evaluate \texttt{EMSCO}'s capabilities, experiments are conducted on three data sets from the UCI Machine Learning Repository \citep{dua17} and one synthetic data set. To gauge performance under a variety of conditions, several confidence thresholds and feature cost schemes are considered.
Experiments in Section \ref{sec:global} empirically evaluate \texttt{EMSCO}'s capacity for global optimization, and experiments in Section \ref{comparisons} compare \texttt{EMSCO} with various classification protocols---both traditional and budgeted approaches. Due to space constraints, the parameter tuning procedure (a simple but computationally expensive ``sweep'' \citep{tuning} procedure) is moved to the Supplement.

\subsection{Data Sets}\label{datasets}
Each data set's features, costs, and respective EA parameters are detailed in this section. The feature costs are assigned with various scaling functions dependent on the assigned \textit{cost class}---an integer value associated with the cost incurred while acquiring the respective feature. Data sets are chosen to represent pathologies (missing data, mixture of categorical and continuous variables, etc.) occurring frequently in real data.
\subsubsection{Heart Failure Clinical Records}
The Heart Failure Clinical Records data set is available on the UCI Machine Learning Repository and was compiled from 299 patients~\citep{chicco2020}. As seen in Table \ref{tab:heartfeatures}, the data set consists of~$12$ pertinent clinical features that are a mix of categorical, integer, and real values. With the guidance of the data set's author, time classes were assigned to each feature representing the acquisition cost. Features that can be acquired without running a test or a follow-up appointment are assigned time class `1', otherwise, the feature is assigned to time class '2'. For this data set, the cost-scaling function is set as $h(T_i) = 10^{T_i}$, where $T_i$ denotes the cost class of the $i$th feature. See Table \ref{tab:heartfeatures} for a description of the features, their cost classes, and the EA parameters used in experiments. Note, we refer to this data set as ``Heart'' for the duration of this paper.
\begin{table}[H]
\begin{small}
\begin{center}
\renewcommand{\arraystretch}{1.0}
\begin{tabular}{ccl}
\textbf{Feature} & \shortstack{\textbf{Cost}} & \multicolumn{1}{l}{\textbf{Description}} \\ \hline \hline
$\mathcal{F}_1$ & 1 & Age of Patient\\ \hline  
$\mathcal{F}_2$ & 2 & Anaemia: \{0,1\}\\ \hline  
$\mathcal{F}_3$ & 2 & High Blood Pressure: \{0,1\}\\ \hline  
$\mathcal{F}_4$ & 2 & Creatinine Phosphokinase Level\\ \hline  
$\mathcal{F}_5$ & 2 & Diabetes: \{0,1\} \\ \hline 
$\mathcal{F}_6$ & 2 & Ejection Fraction \\ \hline
$\mathcal{F}_7$ & 2 & Platelets \\ \hline
$\mathcal{F}_8$ & 1 & Sex: \{0,1\} \\ \hline 
$\mathcal{F}_9$ & 2 & Serum Creatinine \\ \hline 
$\mathcal{F}_{10}$ & 2 & Serum Sodium \\ \hline 
$\mathcal{F}_{11}$ & 1 & Smoking: \{0,1\} \\ \hline 
$\mathcal{F}_{12}$ & 1 & Follow-Up Period \\ \hline 
Target & N/A & Death Event: \{0,1\} \\ \hline
\end{tabular}
\label{tab:import2}  
%Heart Failure Clinical Records Features. \label{tab:heartfeatures}}}
\end{center}
\end{small}
\end{table}
\begin{table}[H] 
\begin{small}
\begin{center}
\renewcommand{\arraystretch}{1.0}
\begin{tabular}{ccl}
\textbf{Parameter} & \shortstack{\textbf{Value}} & \multicolumn{1}{l}{\textbf{Description}} \\ \hline \hline
$\hat{m}$ & 0.075 & Mutation Rate\\ \hline
$\hat{r}$ & 0.75 & Crossover Rate\\ \hline
$b$ & 0.2 & Elite Population Percent\\ \hline
$\Norm{G}$ & 300 & Population Size\\ \hline
$\beta$ & 2 & Mutation Bias Parameter\\ \hline
$\hat{p}$ & 0.75 & Confidence Threshold\\ \hline
\hline 
\end{tabular}
\label{tab:import2}  
\caption{Heart Failure Clinical Records Features and Parameters \label{tab:heartparams}}\label{tab:heartfeatures}
\end{center}
\end{small}
\end{table}

\subsubsection{Pima Diabetes Data Set}
The Pima Diabetes data set \citep{dua17} is a popular data set for evaluating performance of classification models. An array of both integer and real-valued features are used to predict whether or not the patient will be diagnosed with diabetes. To evaluate performance of \texttt{EMSCO} on feature sets with multiple cost classes, values $T_i \in \{1,2,3\}$ were assigned to each feature depending on the time and complications inherent in conducting the individual tests. For this data set, we use the linear scaling function $h(T_i) = 100T_i$. See Table \ref{tab:pimafeatures} for a description of the features and their corresponding cost classes. Note, we refer to this data set as ``Pima'' for the remainder of this paper.
\begin{table}[H]
\begin{small}
\begin{center}
\renewcommand{\arraystretch}{1.0}
\begin{tabular}{ccl}
\textbf{Feature} & \shortstack{\textbf{Cost}} & \multicolumn{1}{l}{\textbf{Description}} \\ \hline \hline
$\mathcal{F}_1$ & 1 & Number of times pregnant\\ \hline  
$\mathcal{F}_2$ & 3 & Plasma glucose concentration\\ \hline  
$\mathcal{F}_3$ & 2 & Diastolic blood pressure\\ \hline  
$\mathcal{F}_4$ & 2 & Triceps skin fold thickness\\ \hline  
$\mathcal{F}_5$ & 3 & 2-Hour serum insulin \\ \hline 
$\mathcal{F}_6$ & 2 & Body mass index \\ \hline
$\mathcal{F}_7$ & 2 & Diabetes pedigree function \\ \hline
$\mathcal{F}_8$ & 1 & Age \\ \hline 
Target & N/A & Diabetes diagnosis: \{0,1\} \\ \hline 
\end{tabular}
\end{center}
\end{small}
\end{table}
\begin{table}[H]
\begin{small}
\begin{center}
\renewcommand{\arraystretch}{1.0}
\begin{tabular}{ccl}
\textbf{Parameter} & \shortstack{\textbf{Value}} & \multicolumn{1}{l}{\textbf{Description}} \\ \hline \hline
$\hat{m}$ & 0.075 & Mutation Rate\\ \hline
$\hat{r}$ & 0.8 & Crossover Rate\\ \hline
$b$ & 0.20 & Elite Population Percent\\ \hline
$\Norm{G}$ & 300 & Population Size\\ \hline
$\beta$ & 2 & Mutation Bias Parameter\\ \hline
$\hat{p}$ & 0.65 & Confidence Threshold\\ \hline
\hline 
\end{tabular}
\label{tab:import2}  
\caption{Pima Diabetes Data Set Features and Parameters\label{tab:pimafeatures}}
\end{center}
\end{small}
\end{table}
\subsubsection{Synthetic30 Data Set} \label{synthetic30}
The Synthetic30 data set was constructed using \texttt{sklearn}'s \texttt{make\_classification()} function \citep{pedregosa2011}. The data set consists of 30 features and 1000 records. To address multi-class settings, three labels are assigned to the records proportionally. 

To mimic scenarios in which a fraction of features are not discriminative for classification \citep{trapeznikov2013}, the parameter \texttt{n\_important} in \texttt{make\_classification()} was set to 25. Gini index \citep{rf2} is used to rank features in terms of importance and thereafter assign cost classes to each feature. Five such cost classes are used, where the most important features are placed in the highest cost class. Let $x_i$ denote the percentile of the $i$th feature's importance in terms of Gini impurity reduction. Cost class, $T_i \in \{1,2,\ldots,5\}$, is defined as follows:

\[ T_i = \begin{cases} 
      1 & 0\leq x_i \leq .2 \\
      2 & .2 < x_i \leq .4 \\
      3 & .4 < x_i \leq .6 \\
      4 & .6 < x_i \leq .8 \\
      5 & .8 < x_i \leq 1.0 \\
   \end{cases}
\]
Final cost of the $i$th feature is a linear function of its cost class: $h(T_i)= 100T_i$. This data set is referred to as ``Synthetic30'' for the rest of the paper.
\begin{table}[H]
\begin{small}
\begin{center}
\renewcommand{\arraystretch}{1.0}
\begin{tabular}{ccl}
\textbf{Parameter} & \shortstack{\textbf{Value}} & \multicolumn{1}{l}{\textbf{Description}} \\ \hline \hline
$\hat{m}$ & 0.05 & Mutation Rate\\ \hline
$\hat{r}$ & 0.8 & Crossover Rate\\ \hline
$b$ & 0.20 & Elite Population Percent\\ \hline
$\Norm{G}$ & 300 & Population Size\\ \hline
$\beta$ & 2.5 & Mutation Bias Parameter\\ \hline
$\hat{p}$ & 0.65 & Confidence Threshold\\ \hline
\hline 
\end{tabular}
\label{tab:import2}  
\caption{Synthetic30 Diabetes Data Set Parameters \label{tab:syntheticparams}}
\end{center}
\end{small}
\end{table}
\subsubsection{Australian Credit Approval Data Set}
This data set is available on the UCI Machine Learning Repository \citep{dua17} and consists of 14 features given in credit applications to a large bank. The target for this data set is binary. Feature descriptions are not provided given the sensitive nature of the data, but the values are designated as continuous or categorical. This data set contains missing values.

Three cost classes are assigned to the features according to their mean decrease in impurity (Gini importance), where features of greater importance are assigned greater cost classes. For this data set, the cost-scaling function is set as $h(T_i) = 10^{T_i}$. This data set is referred to as ``Credit'' in this paper.
\begin{table}[H]
\begin{small}
\begin{center}
\renewcommand{\arraystretch}{1.0}
\begin{tabular}{ccl}
\textbf{Feature} & \shortstack{\textbf{Cost}} & \multicolumn{1}{l}{\textbf{Description}} \\ \hline \hline
$\mathcal{F}_1$ & 2 & Categorical:  \{0,1\}\\ \hline  
$\mathcal{F}_2$ & 2 & Continuous \\ \hline  
$\mathcal{F}_3$ & 2 & Continuous\\ \hline  
$\mathcal{F}_4$ & 1 & Categorical: \{1,2,3\}\\ \hline  
$\mathcal{F}_5$ & 2 & Categorical: \{1$\ldots$14\}\\ \hline 
$\mathcal{F}_6$ & 1 & Categorical: \{1$\ldots$9\} \\ \hline
$\mathcal{F}_7$ & 2 & Continuous \\ \hline
$\mathcal{F}_8$ & 3 & Categorical: \{0,1\} \\ \hline 
$\mathcal{F}_9$ & 1 & Categorical: \{0,1\}\\ \hline 
$\mathcal{F}_{10}$ & 2 & Continuous \\ \hline 
$\mathcal{F}_{11}$ & 1 & Categorical: \{0,1\} \\ \hline 
$\mathcal{F}_{12}$ & 1 & Categorical: \{1,2,3\}\\ \hline 
$\mathcal{F}_{13}$ & 1 & Continuous \\ \hline
$\mathcal{F}_{14}$ & 2 & Continuous \\ \hline 
Target & N/A &  Class \{0,1\} \\ \hline 
\end{tabular}
\label{tab:import3}  
\end{center}
\end{small}
\end{table}
\begin{table}[H]
\begin{small}
\begin{center}
\renewcommand{\arraystretch}{1.0}
\begin{tabular}{ccl}
\textbf{Parameter} & \shortstack{\textbf{Value}} & \multicolumn{1}{l}{\textbf{Description}} \\ \hline \hline
$\hat{m}$ & 0.075 & Mutation Rate\\ \hline
$\hat{r}$ & 0.80 & Crossover Rate\\ \hline
$b$ & 0.20 & Elite Population Percent\\ \hline
$\Norm{G}$ & 300 & Population Size\\ \hline
$\beta$ & 2.5 & Mutation Bias Parameter\\ \hline
$\hat{p}$ & 0.75 & Confidence Threshold\\ \hline
\hline 
\end{tabular}
\label{tab:import2}  
\caption{Australian Credit Data Set Features and Parameters \label{tab:creditfeatures}}
\end{center}
\end{small}
\end{table}
     \subsection{Global Optima Experiment} \label{sec:global}
    In these experiments, \texttt{EMSCO}'s global optimization abilities are tested. The object is to show that \texttt{EMSCO} can quickly find and maintain multiple globally non-dominated solutions in a large fitness landscape determined by sample data.
    
    To establish the true Pareto front for a particular data set in a $k$-stage solution space, a brute-force evaluation of all chromosomes in $\mathcal{S}_{(n,k)}$ is necessary. This endeavor presents a great computational burden and took roughly one month on a 32-core compute server to complete for all data sets. The largest experiment (Australian Credit data set: $n=14,k=4$) required evaluation of over 250 million solutions to establish the global Pareto front. Each solution was evaluated by training its respective $k$ stages (subclassifiers) and recording performance of the corresponding multi-stage system on a validation set that approximates input space $\mathcal{X}$. To reduce runtime, each data set's $2^n$ possible subclassifiers were trained and cached for fast evaluation. Despite such measures, the Synthetic30 data set could not be included in these experiments due to excessive computational expense.
    
    After establishing the global Pareto front for each data set and stage-count ($k=2\ldots4)$, \texttt{EMSCO} was run one-hundred times, and the average count of unique Pareto optimal solutions present in each $h$th elite population ($\Bar{X}_h$) was recorded. Let $\mu_h$ denote the true mean number of unique Pareto optimal solutions in \texttt{EMSCO}'s $h$th elite population. $95\%$ confidence intervals for $\mu_h$ are constructed at each generation and depicted in the second column of Figure \ref{fig:globalexperiment}. Note, due to time-constraints, we conducted experiments with $max\_iter=500$.
    
    In all experiments, \texttt{EMSCO} consistently obtained numerous Pareto optimal solutions while requiring between 30-45 minutes of CPU-time to reach the $500$\textsuperscript{th} generation in our computing environment. Property \ref{thm:pres} is manifest in each chart---all curves are monotonic. Nonetheless, experiments suggest that this property does not preclude a wide breadth of search---in several experiments, a large fraction of the entire global frontier was found and maintained in \texttt{EMSCO}'s elite population by the $500$\textsuperscript{th} generation. Table \ref{tab:frontiers} lists the size of each global frontier. Relative to the size of their respective solution spaces, the frontiers are remarkably small.
    
    Consider, for example, the corresponding search space of the Credit $k=4$ experiment---only 171 of roughly $250$ million chromosomes are globally non-dominated. Nevertheless, within minutes, \texttt{EMSCO} effectively traverses the solution space to identify and maintain over $20\%$ of the global frontier, on average.
    
    Likewise, the Heart ($\mathcal{S}_{(12,4)}$) solution space is quite large, with nearly $15$ million chromosomes; however, its Pareto frontier is very small, with only 33 solutions. In the corresponding experiment, \texttt{EMSCO} again demonstrates its ability to quickly traverse a vast solution space with a sparse Pareto frontier and find/maintain a considerable fraction of unique globally non-dominated solutions---in this case, $80\%$ of the global frontier in less than 500 generations.

\begin{table}[H]
\begin{center}
\renewcommand{\arraystretch}{1.0}
\begin{tabular}{cccc}
\textbf{Data Set} & \shortstack{\textbf{k=2}} & \shortstack{\textbf{k=3}} &  \shortstack{\textbf{k=4}} \\  \hline \hline
Heart & 24 & 28 & 33\\ \hline
Credit & 47 & 141 & 171\\ \hline
Pima & 18 & 42 & 65\\ \hline
\end{tabular}
\label{tab:import2}  
\caption{Global Pareto Frontier Sizes\label{tab:frontiers}}
\end{center}
\end{table}
     \input{global_experiment}

     \subsection{Comparison Against Alternative Budgeted Methods} \label{comparisons}
     These experiments evaluate the performance of $\texttt{EMSCO}$ against several alternatives. The classifier \textit{evaluation} cost is set as zero for all methods and data sets, since evaluation at each stage is negligible for the moderately-sized data sets and cheap classification algorithms used in these experiments.
     
     \textit{Greedy Miser} \citep{xu2012} is a variant of stage-wise regression \citep{friedman2001} with a feature-budgeted loss function. In this method, regression trees are added iteratively to form a cost-effective ensemble classifier. To evaluate this method in our setup and compute coverage rates, we incorporate a reject option before label assignment that discards low-confidence predictions under threshold ($\hat{p}$). Because of this modification, we denote this method as GM$^*$.
     
     \textit{Cost-Sensitive Tree of Classifiers} \citep{xu2014} builds a tree of classifiers optimized for a specific sub-partition of the input space. The aim is to ensure that inputs are classified using only the most pertinent features defined for particular regions of the input space. Doing so reduces unnecessary feature extraction while maintaining accuracy. As with Greedy Miser, we add a reject option to this method that discards predictions below a given threshold. Because of this modification, we denote this method as CSTC$^*$.

     A \textit{Cost-Ordered $T$-Stage Classifier} (CO-T) is considered as a heuristic that utilizes a common cost-ordered perspective for ordering features in budgeted systems \citep{trapeznikov2013,wang2014}. In this setup, a stage is added  for each of the $T$ increasing cost classes. A terminal reject option at stage $T$ is used in this method. We also include a \textit{Cost-and-Coverage-Tuned LASSO} method (CaCT LASSO) that performs variable selection and evaluates all inputs in a single stage prior to applying the terminal reject decision. Cost is then computed as the sum of $C_i$ such that the respective regression coefficients are non-zero. This method is tuned on the validation set for $\lambda \in \{0, 0.1, 0.2, 0.3,\ldots,100\}$ while accounting for conclusive accuracy, cost, and coverage via an aggregate linear function ($g_1 + g_2 + (1 - \frac{g_3^*}{\sum \mathcal{C}_i})$). A similar method for budgeted classification (but without concern for coverage) is addressed as a naive approach in both \citep{xu2014} and \citep{kusner}. Finally, a traditional single-stage classifier with no terminal reject option (100\% coverage) is included to benchmark against out-of-box methods. CO-T, CaCT LASSO, and Single-Stage methods all apply logistic regression at each stage to classify inputs.
     
      We consider Greedy Miser and CSTC because they are among the best-cited budgeted learning approaches, have code available online, and can be easily modified to apply confidence-based reject decisions; However, it is important to note that Greedy Miser and CSTC are not optimized with respect to coverage during the training phase. It is also assumed that these methods are sufficiently well-calibrated \citep{lrcalibration} such that their confidence scores roughly approximate true class probabilities as in logistic regression ---the classification algorithm utilized by \texttt{EMSCO} at each stage. These factors can inflate conclusive accuracy and decrease coverage when the reject protocol is applied (if confidence scores are conservative), but cost is not influenced in either case. To our knowledge, there is not another budgeted method with a terminal reject option considering accuracy, feature acquisition cost, and coverage simultaneously. Current methods must therefore be adapted to fit this paper's setup in experiments. 

     \subsubsection{Performance Evaluation}
     To train, tune, and evaluate methods, a $50$-$25$-$25$ training, validation, and testing split is used for each data set. For a given chromosome $Q$, \texttt{EMSCO} uses the training set to learn its subclassifiers ($\mathscr{C}(Q_j)$) comprising each $j$\textsuperscript{th} stage. Each experiment conducts fifty independent trials---\texttt{EMSCO} is run on the validation set with the parameters listed in Section \ref{datasets} and $max\_iter=150, g=20, k=\max(\lceil\frac{n}{2}\rceil,10)$. At each terminal generation, the chromosome with maximum fitness is stored. When the experiment is finished, these chromosomes are used to compute average (50 samples) $g_1,g_2,g^*_3$ values on the test set, marking \texttt{EMSCO}'s performance. Since inverse cost ($g_3$) is population-dependent, \textit{raw} cost ($g_3^*$ as defined in Section \ref{sec:cost}) is displayed for each method in these experiments.  Note, due to the stochastic nature of evolutionary algorithms, a $95\%$ margin of error ($\epsilon_{EA}$) is included in each chart caption (Figures 8-11). The alternative methods are deterministic, and margin of error is not computed. As done in \citep{xu2012} and \citep{xu2014} GM$^*$ and CSTC$^*$'s parameters were tuned during validation using grid search on a large discretized set. With parameters set, performance of these methods is then recorded on the test set.
  
    \subsubsection{Comparison Experiment Results} \label{sec:compresults}
    \newsavebox{\pimaCon}
\begin{lrbox}{\pimaCon}
\begin{tikzpicture}
\begin{axis} [
   xbar,
   xmin = 0,
   xmax = 105.0, 
   title = Conclusive Classifications for Pima Data,
   xlabel = {Percent Conclusive ($\epsilon_{EA} \leq 0.002$) },
   symbolic y coords={CSTC$^*$,GM$^*$,EMSCO,Single-Stage,CO-T,CaCT LASSO},%,Jenks},
   ytick = data,
   xmajorgrids = true,
   y = 0.7cm,
   x filter/.code={\pgfmathparse{#1*100}\pgfmathresult},
   enlarge y limits = {abs = 0.5cm},
   enlarge x limits = {abs = 0.5cm},
   nodes near coords
   ]
  \addplot[draw = blue, fill=blue!20] coordinates {
     (0.623,CSTC$^*$)
     (0.353,GM$^*$)
     (0.986,EMSCO)
     (1,Single-Stage)
     (.968,CO-T)
     (0.83,CaCT LASSO)
     %(0.916,Jenks)
    };
\end{axis}
\end{tikzpicture}
\end{lrbox}

\newsavebox{\pimaAcc}
\begin{lrbox}{\pimaAcc}
\begin{tikzpicture}
\begin{axis} [
   xbar,
   xmax = 100.0,
   xmin = 0,
   title = Conclusive Accuracy for Pima Data,
   xlabel = {Percent Accuracy ($\epsilon_{EA} \leq 0.0002$)},
   symbolic y coords={CSTC$^*$,GM$^*$,EMSCO,Single-Stage,CO-T,CaCT LASSO},%,Jenks},
   ytick = data,
   xmajorgrids = true,
   y = 0.7cm,
   x filter/.code={\pgfmathparse{#1*100}\pgfmathresult},
   enlarge y limits = {abs = 0.5cm},
   enlarge x limits = {abs = 0.5cm},
   nodes near coords
   ]
  \addplot[draw = red, fill=red!20] coordinates {
%      (0.44,CSTC$^*$)
%      (0.42,Greedy)
     (0.931,CSTC$^*$)
     (0.897,GM$^*$)
     (0.701,EMSCO)
     (0.751,Single-Stage)
     (.72,CO-T)
     (0.818,CaCT LASSO)
     %(0.916,Jenks)
    };
\end{axis}
\end{tikzpicture}
\end{lrbox}

\newsavebox{\pimaCost}
\begin{lrbox}{\pimaCost}
\begin{tikzpicture}
\begin{axis} [
   xbar,
   xmax=1800,
   title = Classification Cost for Pima Data,
   xlabel = {Cost ($\epsilon_{EA} = \leq 1.4$)},
   symbolic y coords={CSTC$^*$,GM$^*$,EMSCO,Single-Stage,CO-T,CaCT LASSO},%,Jenks},
   ytick = data,
   xmajorgrids = true,
   y = 0.7cm,
   enlarge y limits = {abs = 0.5cm},
   enlarge x limits = {abs = 0.5cm},
   nodes near coords
   ]
  \addplot[draw = green, fill=green!20] coordinates {
     (623.7,CSTC$^*$)
     (700.3,GM$^*$)
     (242.0,EMSCO)
     (1600.0,Single-Stage)
     (524,CO-T)
     (1100,CaCT LASSO)
     %(0.916,Jenks)
    };
\end{axis}
\end{tikzpicture}
\end{lrbox}

\begin{figure}
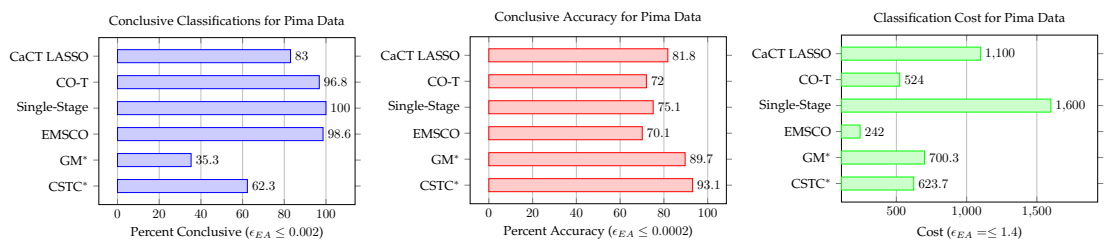

\begin{center}
\resizebox{\textwidth}{!}{% 
\begin{tabular}{ccc}
{\usebox{\pimaCon}} & 
{\usebox{\pimaAcc}} &
{\usebox{\pimaCost}}
\end{tabular}
}
\caption{\small{Comparison results for the Pima Diabetes Data Set.}}
\end{center}
\end{figure}
    \newsavebox{\creditCon}
\begin{lrbox}{\creditCon}
\begin{tikzpicture}
\begin{axis} [
   xbar,
   xmin = 0.0, 
   xmax = 107.0, 
   title = Conclusive Classifications for Credit Data,
   xlabel = {Percent Conclusive ($\epsilon_{EA} \leq 0.006)$},
   symbolic y coords={CSTC$^*$,GM$^*$,EMSCO,Single-Stage,CO-T,CaCT LASSO},%,Jenks},
   ytick = data,
   xmajorgrids = true,
   y = 0.7cm,
   x filter/.code={\pgfmathparse{#1*100}\pgfmathresult},
   enlarge y limits = {abs = 0.5cm},
   enlarge x limits = {abs = 0.5cm},
   nodes near coords
   ]
  \addplot[draw = blue, fill=blue!20] coordinates {
%      (0.14,CSTC)
%      (0.62,GM$^*$)
     (0.775,CSTC$^*$)
     (0.64,GM$^*$)
     (0.961,EMSCO)
     (1,Single-Stage)
     (0.93,CO-T)
     (0.80,CaCT LASSO)
     %(0,Jenks)
    };
\end{axis}
\end{tikzpicture}
\end{lrbox}

\newsavebox{\creditAcc}
\begin{lrbox}{\creditAcc}
\begin{tikzpicture}
\begin{axis} [
   xbar,
   xmin = 0.0, 
   xmax = 104.0, 
   title = Conclusive Accuracy for Credit Data,
   xlabel = {Percent Accuracy ($\epsilon_{EA} \leq 0.004)$},
   symbolic y coords={CSTC$^*$,GM$^*$,EMSCO,Single-Stage,CO-T,CaCT LASSO},%Jenks},
   ytick = data,
   xmajorgrids = true,
   y = 0.7cm,
   x filter/.code={\pgfmathparse{#1*100}\pgfmathresult},
   enlarge y limits = {abs = 0.5cm},
   enlarge x limits = {abs = 0.5cm},
   nodes near coords
   ]
  \addplot[draw = red, fill = red!20] coordinates {
%      (0.85,CSTC$^*$)
%      (0.95,GM$^*$)
     (0.948,CSTC$^*$)
     (0.921,GM$^*$)
     (0.820,EMSCO)
     (0.838,Single-Stage)
     (0.844,CO-T)
     (0.89,CaCT LASSO)
     %(0,Jenks)
    };
\end{axis}
\end{tikzpicture}
\end{lrbox}

\newsavebox{\creditCost}
\begin{lrbox}{\creditCost}
\begin{tikzpicture}
\begin{axis} [
   xbar,
   xmin = 0,
   title = Classification Cost for Credit Data,
   xlabel = {Cost ($\epsilon_{EA} \leq 16.5)$},
   symbolic y coords={CSTC$^*$,GM$^*$,EMSCO,Single-Stage,CO-T,CaCT LASSO},%,Jenks},
   ytick = data,
   xmajorgrids = true,
   y = 0.7cm,
   xmax = 2000,
   enlarge y limits = {abs = 0.5cm},
   enlarge x limits = {upper, abs = 0.5cm},
   nodes near coords
   ]
  \addplot[draw = green, fill = green!20] coordinates {
%      (108.0,CSTC$^*$)
%      (672.0,GM$^*$)
     (492.29,CSTC$^*$)
     (1133.7,GM$^*$)
     (365.889,EMSCO)
     (1760,Single-Stage)
     (695.8,CO-T)
     (1420,CaCT LASSO)
     %(0,Jenks)
    };
\end{axis}
\end{tikzpicture}
\end{lrbox}

\begin{figure}
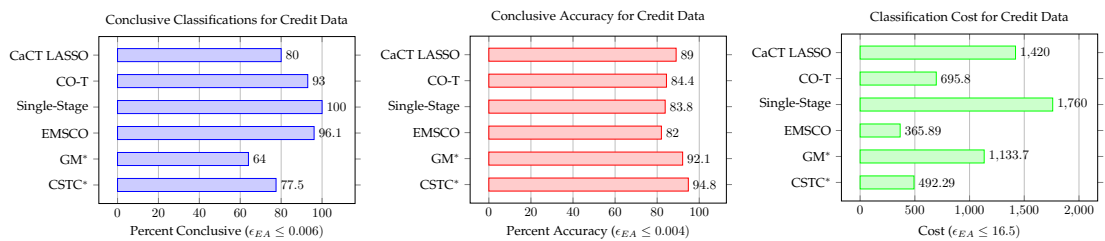

\begin{center}
\resizebox{\textwidth}{!}{% 
\begin{tabular}{ccc}
{\usebox{\creditCon}} & 
{\usebox{\creditAcc}} &
{\usebox{\creditCost}}
\end{tabular}
}
\caption{\small{Comparison results for the Australian Credit Data Set.}}
\end{center}
\end{figure}
    \newsavebox{\heartCon}
\begin{lrbox}{\heartCon}
\begin{tikzpicture}
\begin{axis} [
   xbar,
   xmin = 0.0, 
   xmax = 102.0, 
   title = Conclusive Classifications for Heart Data,
   xlabel = {Percent Conclusive ($\epsilon_{EA} \leq  0.005)$},
   symbolic y coords={CSTC$^*$,GM$^*$,EMSCO,Single-Stage,CO-T,CaCT LASSO},%,Jenks},
   ytick = data,
   xmajorgrids = true,
   y = 0.7cm,
   x filter/.code={\pgfmathparse{#1*100}\pgfmathresult},
   enlarge y limits = {abs = 0.5cm},
   enlarge x limits = {abs = 0.5cm},
   nodes near coords
   ]
  \addplot[draw = blue, fill=blue!20] coordinates {
     (0.552,CSTC$^*$)
     (0.72,GM$^*$)
     (0.89,EMSCO)
     (1.0,Single-Stage)
     (.72,CO-T)
     (0.65,CaCT LASSO)
    };
\end{axis}
\end{tikzpicture}
\end{lrbox}

\newsavebox{\heartAcc}
\begin{lrbox}{\heartAcc}
\begin{tikzpicture}
\begin{axis} [
   xbar,
   xmin = 0.0, 
   xmax = 103.0, 
   xmin = 0,
   title = Conclusive Accuracy for Heart Data,
   xlabel = {Percent Accuracy ($\epsilon_{EA} \leq .001)$},
   symbolic y coords={CSTC$^*$,GM$^*$,EMSCO,Single-Stage,CO-T,CaCT LASSO},%,Jenks},
   ytick = data,
   xmajorgrids = true,
   y = 0.7cm,
   x filter/.code={\pgfmathparse{#1*100}\pgfmathresult},
   enlarge y limits = {abs = 0.5cm},
   enlarge x limits = {abs = 0.5cm},
   nodes near coords
   ]
  \addplot[draw = red, fill = red!20] coordinates {
     (0.971,CSTC$^*$)
     (0.925,GM$^*$)
     (.851,EMSCO)
     (0.840,Single-Stage)
     (.871,CO-T)
     (.877,CaCT LASSO)
    };
\end{axis}
\end{tikzpicture}
\end{lrbox}

\newsavebox{\heartCost}
\begin{lrbox}{\heartCost}
\begin{tikzpicture}
\begin{axis} [
   xbar,
   title = Classification Cost for Heart Data,
   xlabel = {Cost ($\epsilon_{EA} \leq 4.2)$},
   symbolic y coords={CSTC$^*$,GM$^*$,EMSCO,Single-Stage,CO-T,CaCT LASSO},%,Jenks},
   ytick = data,
   xmajorgrids = true,
   y = 0.7cm,
   xmax = 900,
   enlarge y limits = {abs = 0.5cm},
   enlarge x limits = {upper, abs = 0.5cm},
   nodes near coords
   ]
  \addplot[draw = green, fill = green!20] coordinates {
     (47.9,CSTC$^*$)
     (38.6,GM$^*$)
     (102.4,EMSCO)
     (750.0,Single-Stage)
     (255.3,CO-T)
     (120.0,CaCT LASSO)
    };
\end{axis}
\end{tikzpicture}
\end{lrbox}

\newsavebox{\heartFpr}
\begin{lrbox}{\heartFpr}
\begin{tikzpicture}
\begin{axis} [
   xbar,
   title = False Positive Rate for Heart Data,
   xlabel = {FPR ($\epsilon_{EA} \leq .002)$},
   symbolic y coords={CSTC$^*$,GM$^*$,EMSCO,Single-Stage},%,Jenks},
   ytick = data,
   xmajorgrids = true,
   y = 0.7cm,
   xmax = 100,
   x filter/.code={\pgfmathparse{#1*100}\pgfmathresult},
   enlarge y limits = {abs = 0.5cm},
   enlarge x limits = {upper, abs = 0.5cm},
   nodes near coords
   ]
  \addplot[draw = black, fill = black!20] coordinates {
     (0.634,CSTC$^*$)
     (0,GM$^*$)
     (0.044,EMSCO)
     (0.061,Single-Stage)
     %(255.3,Jenks)
    };
\end{axis}
\end{tikzpicture}
\end{lrbox}

\begin{figure}
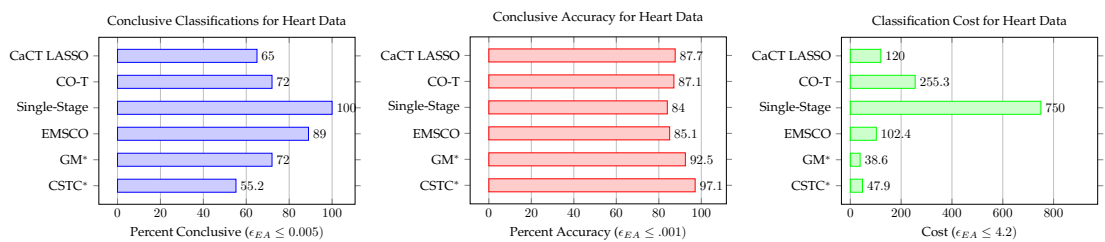

\begin{center}
\resizebox{\textwidth}{!}{% 
\begin{tabular}{ccc}
{\usebox{\heartCon}} & 
{\usebox{\heartAcc}} &
{\usebox{\heartCost}} 
\end{tabular}
}
\caption{\small{Comparison results for the Heart Failure Clinical Records Data Set.}}
\end{center}
\end{figure}
    \newsavebox{\synCon}
\begin{lrbox}{\synCon}
\begin{tikzpicture}
\begin{axis} [
   xbar,
   xmin = 0.0, 
   xmax = 105.0, 
   title = Conclusive Classifications for Synthetic Data,
   xlabel = {Percent Conclusive ($\epsilon_{EA} \leq 0.005)$},
   symbolic y coords={CSTC$^*$,GM$^*$,EMSCO,Single-Stage,CO-T,CaCT LASSO},%,Jenks},
   ytick = data,
   xmajorgrids = true,
   y = 0.7cm,
   x filter/.code={\pgfmathparse{#1*100}\pgfmathresult},
   enlarge y limits = {abs = 0.5cm},
   enlarge x limits = {abs = 0.5cm},
   nodes near coords
   ]
  \addplot[draw = blue, fill=blue!20] coordinates {
     (0.282,CSTC$^*$)
     (0.832,GM$^*$)
     (.928,EMSCO)
     (1,Single-Stage)
     (0.856,CO-T)
     (.64,CaCT LASSO)
     %(0.916,Jenks)
    };
\end{axis}
\end{tikzpicture}
\end{lrbox}

\newsavebox{\synAcc}
\begin{lrbox}{\synAcc}
\begin{tikzpicture}
\begin{axis} [
   xbar,
   xmin = 0, 
   xmax = 104.0, 
   title = Conclusive Accuracy for Synthetic Data ,
   xlabel = {Percent Accuracy ($\epsilon_{EA} \leq 0.01)$},
   symbolic y coords={CSTC$^*$,GM$^*$,EMSCO,Single-Stage,CO-T,CaCT LASSO},%,Jenks},
   ytick = data,
   xmajorgrids = true,
   y = 0.7cm,
   x filter/.code={\pgfmathparse{#1*100}\pgfmathresult},
   enlarge y limits = {abs = 0.5cm},
   enlarge x limits = {abs = 0.5cm},
   nodes near coords
   ]
  \addplot[draw = red, fill=red!20] coordinates {
     (0.422,CSTC$^*$)
     (0.442,GM$^*$)
     (0.67,EMSCO)
     (0.724,Single-Stage)
     (0.6915,CO-T)
     (0.79,CaCT LASSO)
     %(0.916,Jenks)
    };
\end{axis}
\end{tikzpicture}
\end{lrbox}

\newsavebox{\synCost}
\begin{lrbox}{\synCost}
\begin{tikzpicture}
\begin{axis} [
   xbar,
   xmax = 10000, 
   title = Classification Cost for Synthetic Data,
   xlabel = {Cost ($\epsilon_{EA} \leq 65)$},
   symbolic y coords={CSTC$^*$,GM$^*$,EMSCO,Single-Stage,CO-T,CaCT LASSO},%,Jenks},
   ytick = data,
   xmajorgrids = true,
   y = 0.7cm,
   enlarge y limits = {abs = 0.5cm},
   enlarge x limits = {abs = 0.5cm},
   nodes near coords
   ]
  \addplot[draw = green, fill=green!20] coordinates {
     (4060.15,CSTC$^*$)
     (1475.04,GM$^*$)
     (3278.7,EMSCO)
     (8600.0,Single-Stage)
     (6416.8,CO-T)
     (2600,CaCT LASSO)
    };

\end{axis}
\end{tikzpicture}
\end{lrbox}

\begin{figure}
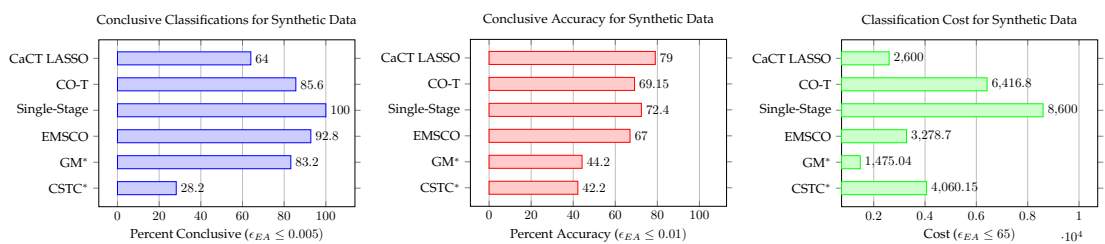

\begin{center}
\resizebox{\textwidth}{!}{% 
\begin{tabular}{ccc}
{\usebox{\synCon}} & 
{\usebox{\synAcc}} &
{\usebox{\synCost}} 
\end{tabular}
}
\caption{\small{Comparison results for the Synthetic30 Data Set.}}
\end{center}
\end{figure} 
    \begin{itemize}
        \item \textit{Pima Data Set}. \texttt{EMSCO} exhibits strong performance with lowest cost (80\% lower than a traditional single stage classifier), and highest coverage among all budgeted methods. Although \texttt{EMSCO}'s accuracy on this data set is similar to a single-stage classifier, it is lower than the other budgeted methods'. However, in our scope, since a $65\%$ confidence threshold was applied, conclusive accuracy of $70.1\%$ is considered sufficient given the system monitor's expectations. In comparison, CSTC$^*$ and GM$^*$ yielded a higher accuracy that may be preferable in certain instances if low conclusiveness and higher cost are acceptable. CO-T provided high conclusiveness and sufficient accuracy; However, this method has twice \texttt{EMSCO}'s cost, lower conclusiveness, and only marginally greater accuracy (less than 2\%). CaCT LASSO did not effectively reduce cost compared to the opposing budgeted methods.

        \item \textit{Credit Data Set}. For the Credit dataset, \texttt{EMSCO} again reported the lowest cost among all tested methods (79\% lower than a traditional single-stage classifier), and highest conclusiveness among budgeted protocols. With respect to accuracy, \texttt{EMSCO} offered good performance ($82\%$). GreedyMiser showed relatively low conclusiveness and highly inflated cost. As in the Pima experiment, CSTC$^*$ had highest conclusive accuracy but low conclusiveness compared to alternatives. CO-T achieves high accuracy and conclusiveness in this experiment but yields high cost that is possibly disqualifying in a budgeted setting. Overall, \texttt{EMSCO} and CSTC$^*$ seem to provide the strongest performance on this data set, with \texttt{EMSCO} offering markedly greater coverage but lower conclusive accuracy.

        \item \textit{Heart Data Set}. GM$^*$ stands out in the Heart comparison experiment for its strong performance across all objectives and minimal cost (81\% less than a single-stage classifier and 65\% less than CSTC$^*$). However, conclusiveness is slightly low, ranking ahead of only one other method (CSTC$^*$). CSTC$^*$ likewise provides remarkably high accuracy but only classifies about half of inputs. \texttt{EMSCO} provides strong accuracy ($85.1\%$), the highest conclusiveness among budgeted approaches, and exhibits a significant reduction in cost compared to CO-T and traditional out-of-box logistic regression. CaCT LASSO yields fairly low cost and high accuracy but is second-to-last in terms of coverage. 
    
        \item \textit{Synthetic30 Data Set}. In the Synthetic30 experiment, \texttt{EMSCO} provides balanced performance with highest coverage among the selective methods with terminal reject options. Furthermore, the cost for \texttt{EMSCO} is 62\% less than a single-stage classifier. Note that \texttt{EMSCO} Pareto dominates CSTC$^*$ in this multi-class setting. GM$^*$ offers very low cost and high conclusiveness, but its conclusive accuracy ($44.2\%$) is likely disqualifying given the system monitor's desire to be correct in roughly $65\%$ of instances. CO-T offers high accuracy and coverage but does not successfully reduce cost compared to a traditional out-of-box method (Single-Stage). CaCT LASSO performs quite well overall on this data set with second lowest cost and highest accuracy but terminally rejects a large fraction of inputs ($35\%$).
        
    \end{itemize}
    In all experiments, \texttt{EMSCO} is non-dominated and achieves well-balanced performance across objectives. \texttt{EMSCO}'s conclusive accuracy exceeded $\hat{p}$ in all cases, suggesting \texttt{EMSCO}'s confidence-based reject protocol was effective in maintaining accuracy at or above a specified minimum. In this paper's setup, $g_2 \geq \hat{p}$ can be viewed as sufficient given the preferences of a risk-averse monitor (Section \ref{sec:setup}) applying the threshold, and a monitor with different preferences/misclassifcation penalties could adjust the confidence threshold to increase $g_2$.

    In 2 of 4 total experiments, \texttt{EMSCO} yielded the lowest average classification cost, and in the other 2 experiments was only outperformed in this metric by methods with markedly lower conclusiveness (e.g, Heart-CSTC$^*$/GM$^*$) or low accuracy (e.g., Synthetic30-GM). With regard to conclusiveness, \texttt{EMSCO} assigned labels to, on average, $93.4\%$ of inputs---representing a utility loss of only $6.6\%$ compared to a traditional non-selective method without a terminal reject option. \texttt{EMSCO} likewise offers a $475\%$ average speed increase and an accuracy decrease of only $2.72\%$ compared to a traditional approach that acquires all features for all inputs. GreedyMiser and CSTC$^*$ reported an average utility loss of $45.2\%$ and $36.4\%$, respectively. Note that \texttt{EMSCO}'s accuracy is comparable to the traditional logistic regression model in all experiments, and accuracy could possibly be increased by incorporating a more powerful classification algorithm at each stage. Though, deploying an expensive classification algorithm at each stage could yield non-trivial classifier evaluation cost.
    
    Figure 12 depicts an aggregate performance metric for the methods with respect to $g_1,g_2$ and a global form of inverse cost.  Though this paper focuses primarily on Pareto efficiency, experiments suggest that \texttt{EMSCO} likewise yields strong combined performance according to a straight-forward linear scalarization ($g_1 + g_2 + (1 - \frac{g_3^*}{\sum \mathcal{C}_i})$)  as well.
    \newsavebox{\pimaboxtwo}
\begin{lrbox}{\pimaboxtwo}
\begin{tikzpicture}
\begin{axis} [xbar,xmax=2.75,
 symbolic y coords={CSTC$^*$,GM$^*$,EMSCO,Single-Stage,CO-T,CaCT LASSO},title={Pima},xlabel={$\epsilon_{EA} \leq .005$},
height=7cm,
width=9cm,
ytick=data,
enlarge x limits = {upper, abs = 0.5cm},
xmajorgrids = true,
nodes near coords
 ]
\addplot[draw = gray, fill = yellow!50] coordinates {
     (2.16,CSTC$^*$)
     (1.8,GM$^*$)
     (2.5,EMSCO)
     (1.75,Single-Stage)
     (2.36,CO-T)
     (1.96,CaCT LASSO)
};
%\addplot[draw = red, fill = red!20] coordinates {
%     (7,CSTC$^*$)
%     (9,GM$^*$)
%     (57,EMSCO)
%     (48,Single-Stage)
%     (64,$n$-Stage)
%};
\end{axis}
\end{tikzpicture}
\end{lrbox}

\newsavebox{\creditboxtwo}
\begin{lrbox}{\creditboxtwo}
\begin{tikzpicture}
\begin{axis} [xbar,xmax=2.8,
 symbolic y coords={CSTC$^*$,GM$^*$,EMSCO,Single-Stage,CO-T,CaCT LASSO},title={Credit},xlabel={$\epsilon_{EA} \leq  .01$},
xmajorgrids = true,
height=7cm,
width=9cm,
ytick=data,
nodes near coords
 ]
\addplot[draw = gray, fill = yellow!50] coordinates {
     (2.43,CSTC$^*$)
     (1.89,GM$^*$)
     (2.56,EMSCO)
     (1.83,Single-Stage)
     (2.36,CO-T)
     (1.85,CaCT LASSO)
};
%\addplot[draw = red, fill = red!20] coordinates {
%     (9,CSTC$^*$)
%     (7,GM$^*$)
%     (30,EMSCO)
%     (28,Single-Stage)
%     (37,$n$-Stage)
%};
  \end{axis}
\end{tikzpicture}
\end{lrbox}

\newsavebox{\heartboxtwo}
\begin{lrbox}{\heartboxtwo}
\begin{tikzpicture}
\begin{axis} [xbar,xmax=2.8,
 symbolic y coords={CSTC$^*$,GM$^*$,EMSCO,Single-Stage,CO-T, CaCT LASSO},title={Heart}, xlabel={$\epsilon_{EA} \leq  .013$},
height=7cm,
width=9cm,
ytick=data,
xmajorgrids = true,
nodes near coords
 ]
\addplot[draw = gray, fill = yellow!50] coordinates {
     (2.45,CSTC$^*$)
     (2.59,GM$^*$)
     (2.605,EMSCO)
     (1.84,Single-Stage)
     (2.25,CO-T)
     (2.36,CaCT LASSO)
};
%\addplot[draw = red, fill = red!20] coordinates {
%     (2,CSTC$^*$)
%     (4,GM$^*$)
%     (13,EMSCO)
%     (12,Single-Stage)
%     (13,$n$-Stage)
%};
  \end{axis}
\end{tikzpicture}

\end{lrbox}

\newsavebox{\syntheticboxtwo}
\begin{lrbox}{\syntheticboxtwo}
\begin{tikzpicture}
\begin{axis} [xbar,xmax=2.5,
 symbolic y coords={CSTC$^*$,GM$^*$,EMSCO,Single-Stage,CO-T,CaCT LASSO},title={Synthetic30},
 xlabel={$\epsilon_{EA} \leq .02$},
height=7cm,
width=9cm,
ytick=data,
xmajorgrids = true,
nodes near coords
 ]
\addplot[draw = gray, fill = yellow!50] coordinates {
     (1.23,CSTC$^*$)
     (2.102,GM$^*$)
     (2.21,EMSCO)
     (1.72,Single-Stage)
     (1.8,CO-T)
     (2.127,CaCT LASSO)
};
%\addplot[draw = red, fill = red!20] coordinates {
%     (41,CSTC$^*$)
%     (117,GM$^*$)
%     (77,EMSCO)
%     (69,Single-Stage)
%     (90,$n$-Stage)
%};
  \end{axis}
\end{tikzpicture}
\end{lrbox}

\begin{figure}
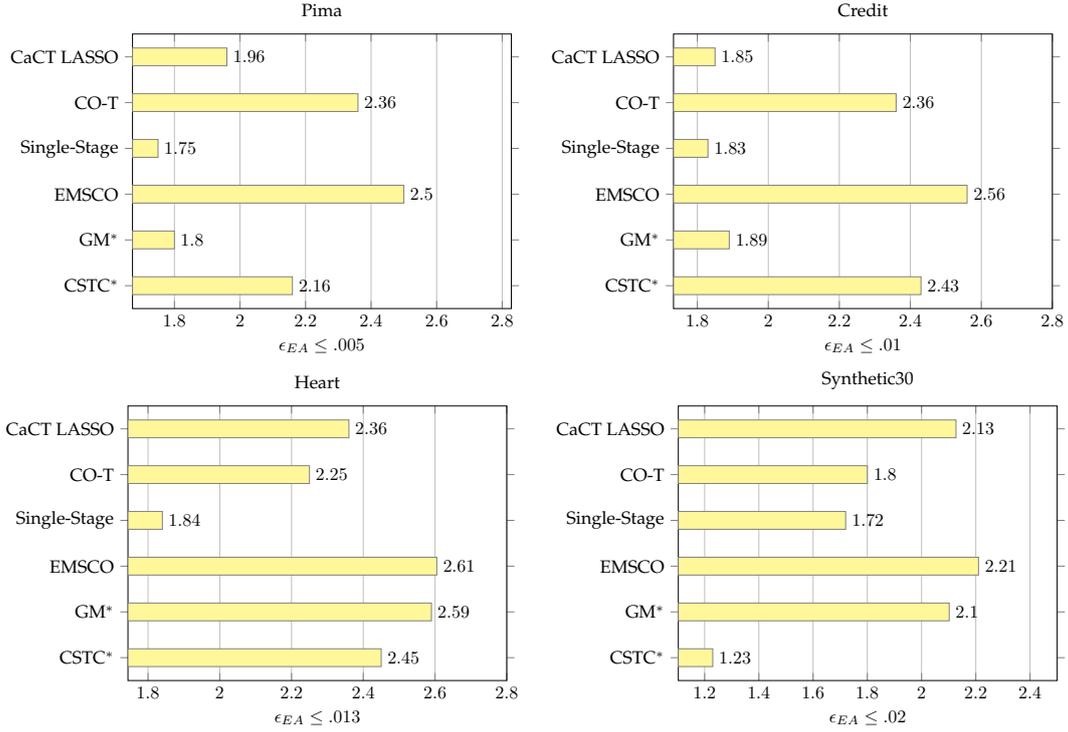

\begin{center}
\resizebox{\textwidth}{!}{% 
\begin{tabular}{cccc}
{\usebox{\pimaboxtwo}} & 
{\usebox{\creditboxtwo}}&\\
{\usebox{\heartboxtwo}}&
{\usebox{\syntheticboxtwo}}
\end{tabular}
}
\caption{\small{\textit{Combined performance}. The sum $g_1 + g_2 + (1 - \frac{g_3^*}{\sum \mathcal{C}_i})$ is computed as a global aggregate measure of performance. \texttt{EMSCO}'s average (50 samples) performance according to this metric is displayed.}}
\end{center}
\label{fig:weighted_sum}
\end{figure}
    \subsubsection{Analysis of \texttt{EMSCO}'s Solutions}
    There is not a broad pattern governing \texttt{EMSCO}'s optimization behaviour; however, an analysis of its returned solutions offers some insight. See Figure 13 for a visual depiction of the returned solutions' average stage assignments with respect to cost classes. For each $i$\textsuperscript{th} comparison experiment, we record the chromosome representation $[Q]_{max_i}$ of the top-scoring solution in the terminal generation according to fitness. Once the experiment is finished, there are fifty such chromosome representations, and we take the element-wise mean of these vectors ($\Bar{[Q]}_{max}$) to assess patterns in stage assignment. In Figure 13, we display the average stage assignment for each feature in $\Bar{[Q]}_{max}$ grouped according to cost class.
    One notable feature of this data involves the poor association between cost class and average stage assignment. Indeed, features in the greatest cost classes are commonly placed in early stages. While a cost-ordered assignment protocol is intuitive, an exhaustive search with \texttt{EMSCO} suggests that assigning a small number of informative/expensive features to early stages is advantageous in certain contexts (See the Synthetic30 and Pima plots in Figure 13). \texttt{EMSCO} also appears to be combining expensive and inexpensive features in single stages (For example, see the Synthetic30 plot in Figure 13---the first two cost classes overlap markedly). This is possibly due to feature interactions in which an inexpensive feature becomes highly informative when grouped with a feature in a greater cost class (or vice versa). The benefits of such interactions may outweigh the resulting increased stage cost.
    
    \newsavebox{\pimabox}
\begin{lrbox}{\pimabox}
\begin{tikzpicture}
\begin{axis}[grid style={line width=.15pt, draw=gray!30},xmin=0,xmax=4,ymax=5,ymin=0,major grid style={line width=.1pt,draw=gray!50},title={Pima},ylabel={Avg. Stage Assn.},xlabel=$T_i$,legend pos=north west,grid=both,
    minor x tick num={4},
    minor y tick num={4},
    minor z tick num={4},
    enlargelimits={abs=0},
    ticklabel style={font=\tiny,fill=white},
    axis line style={latex-latex}]
 
%plot in all axes

%lot accuracy vs. cost (x= g_2 vs. z = g_3)
\addplot[mark=*, only marks, color=red!70] table [x = col1, y = col2] 
{
col1 col2
1 2.08
3 1.06
2 3.0
2 2.98
3 3.0
2 1.94
2 2.98
1 0.04
};

%\legend{One-Step Neighborhood,2D $g_2$-$g_3$ trade-off}
\end{axis}
\end{tikzpicture}
\end{lrbox}

\newsavebox{\creditbox}
\begin{lrbox}{\creditbox}
\begin{tikzpicture}
\begin{axis}[grid style={line width=.15pt, draw=gray!30},xmin=0,xmax=4,ymax=5,ymin=0,major grid style={line width=.1pt,draw=gray!50},title={Credit},ylabel={Avg. Stage Assn.},xlabel=$T_i$,legend pos=north west,grid=both,
    minor x tick num={4},
    minor y tick num={4},
    minor z tick num={4},
    enlargelimits={abs=0},
    ticklabel style={font=\tiny,fill=white},
    axis line style={latex-latex}]
 
%plot in all axes

%lot accuracy vs. cost (x= g_2 vs. z = g_3)
\addplot[mark=*, only marks, color=red!70] table [x = col1, y = col2] 
{
col1 col2
2 4.28
2 3.94
2 4.08
1 2.28
2 1.16
1. 0.12
2 3.9
3 4.04
1 0.0
2 2.82
1 2.58
1 2.98
1 2.38
2 2.78
};

%\legend{One-Step Neighborhood,2D $g_2$-$g_3$ trade-off}
\end{axis}
\end{tikzpicture}
\end{lrbox}

\newsavebox{\heartbox}
\begin{lrbox}{\heartbox}
\begin{tikzpicture}
\begin{axis}[grid style={line width=.15pt, draw=gray!30},xmin=0,xmax=3,ymax=5,ymin=0,major grid style={line width=.1pt,draw=gray!50},title={Heart},ylabel={Avg. Stage Assn.},xlabel=$T_i$,legend pos=north west,grid=both,
    minor x tick num={4},
    minor y tick num={4},
    minor z tick num={4},
    enlargelimits={abs=0},
    ticklabel style={font=\tiny,fill=white},
    axis line style={latex-latex}]
 
%plot in all axes

%lot accuracy vs. cost (x= g_2 vs. z = g_3)
\addplot[mark=*, only marks, color=red!70] table [x = col1, y = col2] 
{
col1 col2
1 2.5
2 3.6333333333333333
2 4.333333333333333
2 4.833333333333333
1 0.9333333333333333
2 4.633333333333334
2 4.5
1 0.0
2 4.433333333333334
2 4.433333333333334
1 3.8
1 2.2666666666666666
};

%\legend{One-Step Neighborhood,2D $g_2$-$g_3$ trade-off}
\end{axis}
\end{tikzpicture}
\end{lrbox}

\newsavebox{\syntheticbox}
\begin{lrbox}{\syntheticbox}
\begin{tikzpicture}
\begin{axis}[grid style={line width=.15pt, draw=gray!30},xmin=0,xmax=5.5,ymax=10,ymin=0,major grid style={line width=.1pt,draw=gray!50},title={Synthetic30},ylabel={Avg. Stage Assn.},xlabel={$T_i$},legend pos=north west,grid=both,
    minor x tick num={4},
    minor y tick num={4},
    minor z tick num={4},
    enlargelimits={abs=0},
    ticklabel style={font=\tiny,fill=white},
    axis line style={latex-latex}]
 
%plot in all axes

%lot accuracy vs. cost (x= g_2 vs. z = g_3)
\addplot[mark=*, only marks, color=red!70] table [x = col1, y = col2] 
{
col1 col2
4 2.36
3 5.64
1 6.36
2 6.42
5 0.54
1 5.86
2 6.54
3 5.28
3 3.28
1 4.92
1 6.62
2 4.58
5 6.28
1 6.62
2 6.86
5 2.66
4 5.96
3 6.78
4 6.2
3 6.04
1 5.66
2 5.26
5 2.42
5 4.32
4 1.92
3 7.46
4 0.38
4 5.06
1 4.8
2 4.9
};

%\legend{One-Step Neighborhood,2D $g_2$-$g_3$ trade-off}
\end{axis}
\end{tikzpicture}
\end{lrbox}

\begin{figure}
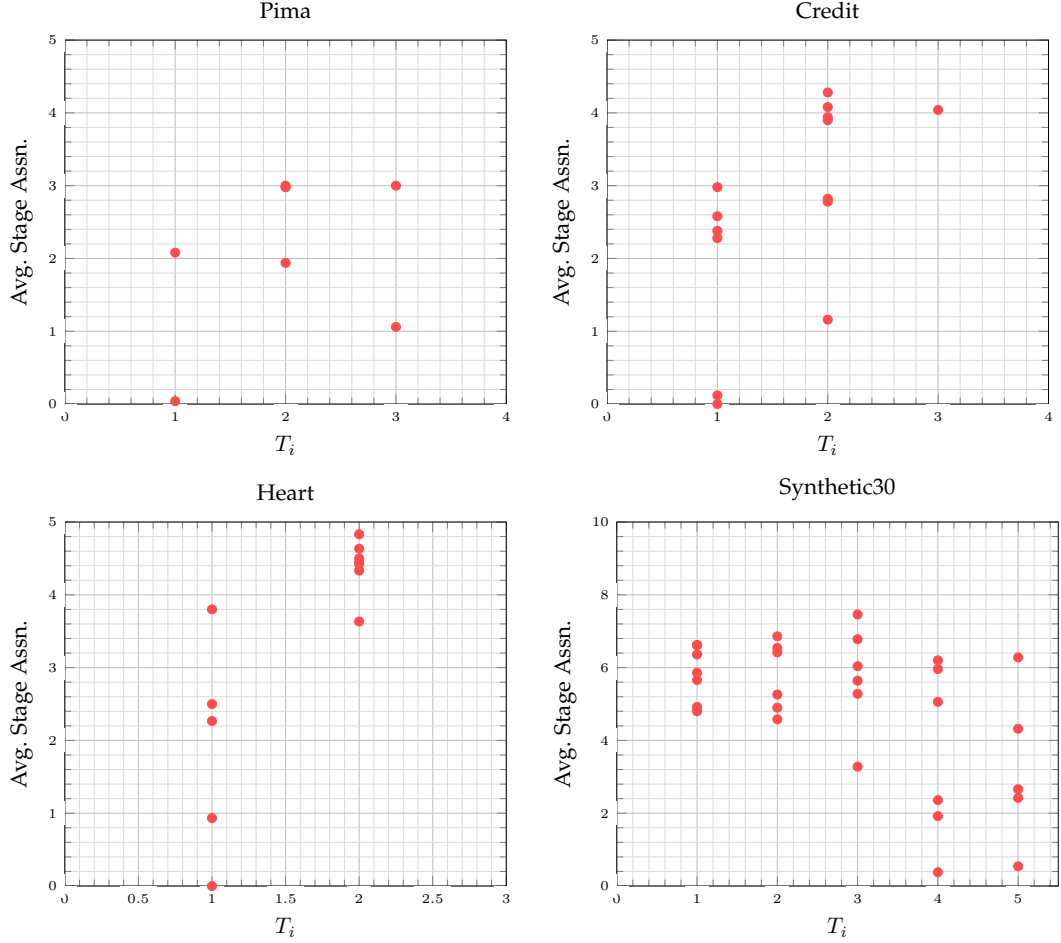

\begin{center}
\resizebox{\textwidth}{!}{% 
\begin{tabular}{cccc}
{\usebox{\pimabox}} & 
{\usebox{\creditbox}}&\\
{\usebox{\heartbox}}&
{\usebox{\syntheticbox}}
\end{tabular}
}
\caption{\small{\textit{Average stage assignment
for each feature in $[\Bar{Q}]_{max}$ grouped according to cost class}.}}
\end{center}
\end{figure}
    \section{Conclusion}
    In this manuscript, \texttt{EMSCO}, an evolutionary algorithm for optimizing high-coverage, low-cost multi-stage classifiers, was proposed. This method was shown capable of effectively managing three important objectives and providing balanced performance. Compared to alternative approaches, \texttt{EMSCO} offered solutions with high coverage, low cost, and sufficient accuracy. Additionally, the method proved itself capable of identifying globally Pareto optimal solutions within very large solution spaces in multiple experimental settings. However, there are several limitations of \texttt{EMSCO} that should be noted.
    
    Due to their heuristic nature, evolutionary algorithms do not provide guarantees of optimality. While the experiments in Figure \ref{fig:globalexperiment} are encouraging, similar results cannot be ensured in the general case. It is particularly important to consider whether these patterns persist in the context of higher dimensional data, but the combinatorial burden imposed by our setup makes addressing this case difficult for very large $n$. Another downside to a population-based approach such as \texttt{EMSCO} involves training/validation time. Indeed, if $\mathcal{S}_{(n,k)}$ is very large (trillions of solutions), \texttt{EMSCO} could require many generations to settle on a strong solution.
    \subsection{Future Work}
    \textit{Feature Removal Option}. To further reduce cost, future work may benefit from including a \textit{feature removal option} in the chromosome representation. For example, the mutation operator could be modified to allow assignment $[Q]_i = -1$, representing removal of feature $i$ from all stages. This protocol may prove particularly useful when $n$ is large and feature selection can reduce complexity and cost \citep{yu2020}.

    \textit{New Objectives}. Currently, \texttt{EMSCO} optimizes with respect to three particular objectives, but these could be altered depending on the setting. For example, certain contexts impose severe penalties for false positives that may warrant explicit consideration of this value as an objective. In light of this, one potentially useful objective is inverse false-positive rate (FPR)---since \texttt{EMSCO} maximizes objectives in $[0,1]$, the value $1 - $FPR$(Q)$ could be included as an objective.  Additionally, to bias the method towards low stage counts, an inverse stage count could be leveraged as an objective. This modification could replace or supplement the bias introduced during mutation (Section \ref{sec:mutation}).
        
    \textit{Enhanced Reject Decision}. The current reject protocol (confidence thresholds) yields desired performance in experiments, is fast to compute, and intuitive; However, it is perhaps too simplistic for certain settings. Given that many real-world scenarios impose varying costs for false positives and false negatives \citep{threshold}, future work may benefit from incorporating different confidence thresholds for positive and negative predictions. Likewise, for high $k$, where there are additional opportunities for marginal, incorrect predictions to be accepted, the reject decision could be modified in an attempt to preserve accuracy at or above the initial specified level ($\hat{p}$) with higher probability.

\bibliographystyle{apalike}
\bibliography{ecjsample}

\begin{thebibliography}{}

\bibitem[Artières et~al., 2016]{block}
Artières, T., Contardo, G., and Denoyer, L. (2016).
\newblock Recurrent neural networks for adaptive feature acquisition.
\newblock {\em 23rd International Conference on Neural Information Processing}.

\bibitem[Boyd and Vandenberghe, 2004]{boyd}
Boyd, S. and Vandenberghe, L. (2004).
\newblock {\em Convex Optimization}.
\newblock Cambridge University Press.

\bibitem[Casella and Berger, 2001]{casella01}
Casella, G. and Berger, R. (2001).
\newblock {\em Statistical Inference}.
\newblock {Duxbury Resource Center}.

\bibitem[Chicco and Jurman, 2020]{chicco2020}
Chicco, D. and Jurman, G. (2020).
\newblock Heart failure clinical records data set.
\newblock UCI Machine Learning Repository.

\bibitem[Cortes et~al., 2016]{cortes}
Cortes, C., DeSalvo, G., and Mohri, M. (2016).
\newblock Learning with rejection.

\bibitem[De~Jong, 2007]{tuning}
De~Jong, K. (2007).
\newblock Parameter setting in eas: a 30 year perspective.
\newblock In {\em Parameter setting in evolutionary algorithms}, pages 1--18.
  Springer.

\bibitem[Deb, 2001]{deb01}
Deb, K. (2001).
\newblock {\em Multi-Objective Optimization Using Evolutionary Algorithms}.
\newblock John Wiley \& Sons, Inc., New York, NY, USA.

\bibitem[Deb et~al., 2002]{996017}
Deb, K., Pratap, A., Agarwal, S., and Meyarivan, T. (2002).
\newblock A fast and elitist multiobjective genetic algorithm: Nsga-ii.
\newblock {\em IEEE Transactions on Evolutionary Computation}, 6(2):182--197.

\bibitem[Dua and Graff, 2017]{dua17}
Dua, D. and Graff, C. (2017).
\newblock {UCI} machine learning repository.

\bibitem[Dundar and Bi, 2007]{dundar07}
Dundar, M.~M. and Bi, J. (2007).
\newblock Joint optimization of cascaded classifiers for computer aided
  detection.
\newblock In {\em 2007 IEEE Conference on Computer Vision and Pattern
  Recognition}, pages 1--8.

\bibitem[Eiben and Smith, 2015]{eiben}
Eiben, A. and Smith, J. (2015).
\newblock {\em Introduction to Evolutionary Computing}.
\newblock Natural Computing Series. Springer.
\newblock Gebeurtenis: 2nd edition.

\bibitem[El-Yaniv and Wiener, 2010]{yaniv10}
El-Yaniv, R. and Wiener, Y. (2010).
\newblock On the foundations of noise-free selective classification.
\newblock {\em Journal of Machine Learning Research}, 11(53):1605--1641.

\bibitem[Friedman, 2001]{friedman2001}
Friedman, J.~H. (2001).
\newblock {Greedy function approximation: A gradient boosting machine.}
\newblock {\em The Annals of Statistics}, 29(5):1189 -- 1232.

\bibitem[Geifman and El-Yaniv, 2017]{geifman2017}
Geifman, Y. and El-Yaniv, R. (2017).
\newblock Selective classification for deep neural networks.

\bibitem[Hamilton and Fulp, 2020]{hamilton2020}
Hamilton, N.~H. and Fulp, E.~W. (2020).
\newblock An evolutionary approach for constructing multi-stage classifiers.
\newblock In {\em Proceedings of the 2020 Genetic and Evolutionary Computation
  Conference Companion}, GECCO '20, page 1730–1738, New York, NY, USA.
  Association for Computing Machinery.

\bibitem[Hamilton et~al., 2020]{hamilton20202}
Hamilton, N.~H., McKinney, S., Allan, E., and Fulp, E.~W. (2020).
\newblock An efficient multi-stage approach for identifying domain shadowing.
\newblock In {\em ICC 2020 - 2020 IEEE International Conference on
  Communications (ICC)}, pages 1--7.

\bibitem[Janisch et~al., 2019]{jansich}
Janisch, J., Pevn{\'{y}}, T., and Lis{\'{y}}, V. (2019).
\newblock Classification with costly features as a sequential decision-making
  problem.
\newblock {\em CoRR}, abs/1909.02564.

\bibitem[Ji and Carin, 2007]{ji2007}
Ji, S. and Carin, L. (2007).
\newblock Cost-sensitive feature acquisition and classification.
\newblock {\em Pattern Recogn.}, 40(5):1474–1485.

\bibitem[Knuth, 1998]{knuth1998}
Knuth, D.~E. (1998).
\newblock {\em The Art of Computer Programming, Volume 3: (2nd Ed.) Sorting and
  Searching}.
\newblock Addison Wesley Longman Publishing Co., Inc., USA.

\bibitem[Kuhn et~al., 2013]{kuhn2013}
Kuhn, M., Johnson, K., et~al. (2013).
\newblock {\em Applied predictive modeling}, volume~26.
\newblock Springer.

\bibitem[Kull et~al., 2017]{lrcalibration}
Kull, M., Filho, T. M.~S., and Flach, P. (2017).
\newblock {Beyond sigmoids: How to obtain well-calibrated probabilities from
  binary classifiers with beta calibration}.
\newblock {\em Electronic Journal of Statistics}, 11(2):5052 -- 5080.

\bibitem[Kusner et~al., 2014]{kusner}
Kusner, M.~J., Chen, W., Zhou, Q., Xu, Z., Weinberger, K.~Q., and Chen, Y.
  (2014).
\newblock Feature-cost sensitive learning with submodular trees of classifiers.
\newblock In {\em Proceedings of the Twenty-Eighth AAAI Conference on
  Artificial Intelligence}, AAAI'14, page 1939–1945. AAAI Press.

\bibitem[Lichtblau and Stoean, 2019]{cancerconf}
Lichtblau, D. and Stoean, C. (2019).
\newblock Cancer diagnosis through a tandem of classifiers for digitized
  histopathological slides.
\newblock {\em PLoS One}.

\bibitem[Nan et~al., 2016]{nan16}
Nan, F., Wang, J., and Saligrama, V. (2016).
\newblock Pruning random forests for prediction on a budget.
\newblock In {\em Advances in Neural Information Processing Systems 29: Annual
  Conference on Neural Information Processing Systems 2016, December 5-10,
  2016, Barcelona, Spain}, pages 2334--2342.

\bibitem[Nogueira et~al., 2019]{nogueira2019}
Nogueira, R., Yang, W., Cho, K., and Lin, J. (2019).
\newblock Multi-stage document ranking with bert.

\bibitem[Patel et~al., 2019]{patel}
Patel, B.~N., Rosenberg, L.~B., Willcox, G., Baltaxe, D., Lyons, M., Irvin,
  J.~A., Rajpurkar, P., Amrhein, T.~J., Gupta, R., Halabi, S.~S., Langlotz, C.,
  Lo, E., Mammarappallil, J.~G., Mariano, A.~J., Riley, G., Seekins, J., Shen,
  L., Zucker, E., and Lungren, M.~P. (2019).
\newblock Human–machine partnership with artificial intelligence for chest
  radiograph diagnosis.
\newblock {\em NPJ Digital Medicine}, 2.

\bibitem[Pedregosa et~al., 2011]{pedregosa2011}
Pedregosa, F., Varoquaux, G., Gramfort, A., Michel, V., Thirion, B., Grisel,
  O., Blondel, M., Prettenhofer, P., Weiss, R., Dubourg, V., et~al. (2011).
\newblock Scikit-learn: Machine learning in python.
\newblock {\em Journal of machine learning research}, 12(Oct):2825--2830.

\bibitem[Peter et~al., 2017]{peter17}
Peter, S., Diego, F., Hamprecht, F.~A., and Nadler, B. (2017).
\newblock Cost efficient gradient boosting.
\newblock In Guyon, I., Luxburg, U.~V., Bengio, S., Wallach, H., Fergus, R.,
  Vishwanathan, S., and Garnett, R., editors, {\em Advances in Neural
  Information Processing Systems}, volume~30. Curran Associates, Inc.

\bibitem[Scornet, 2020]{rf2}
Scornet, E. (2020).
\newblock Trees, forests, and impurity-based variable importance.

\bibitem[Trapeznikov et~al., 2013]{trapeznikov2013}
Trapeznikov, K., Saligrama, V., and Casta{\~{n}}{\'{o}}n, D. (2013).
\newblock Multi-stage classifier design.
\newblock {\em Machine Learning}, 92(2-3):479--502.

\bibitem[Viola and Jones, 2001]{viola}
Viola, P.~A. and Jones, M.~J. (2001).
\newblock Rapid object detection using a boosted cascade of simple features.
\newblock In {\em CVPR (1)}, pages 511--518. IEEE Computer Society.

\bibitem[Wang et~al., 2014]{wang2014}
Wang, J., Trapeznikov, K., and Saligrama, V. (2014).
\newblock {An LP for Sequential Learning Under Budgets}.
\newblock volume~33 of {\em Proceedings of Machine Learning Research}, pages
  987--995, Reykjavik, Iceland. PMLR.

\bibitem[Weisstein, 2012]{wolfram1}
Weisstein, E.~W. (2012).
\newblock von neumann neighborhood. {From MathWorld---A Wolfram Web Resource}.

\bibitem[Xu et~al., 2012]{xu2012}
Xu, Z., Weinberger, K., and Chapelle, O. (2012).
\newblock The greedy miser: Learning under test-time budgets.

\bibitem[Xu et~al., 2014]{xu2014}
Xu, Z.~E., Kusner, M.~J., Weinberger, K.~Q., Chen, M., and Chapelle, O. (2014).
\newblock Classifier cascades and trees for minimizing feature evaluation cost.
\newblock {\em Journal of Machine Learning Research}, 15(62):2113--2144.

\bibitem[Yu et~al., 2020]{yu2020}
Yu, G., Witten, D., and Bien, J. (2020).
\newblock Controlling costs: Feature selection on a budget.

\bibitem[Zhou and Liu, 2006]{threshold}
Zhou, Z.-H. and Liu, X.-Y. (2006).
\newblock Training cost-sensitive neural networks with methods addressing the
  class imbalance problem.
\newblock {\em IEEE Transactions on Knowledge and Data Engineering},
  18(1):63--77.

\end{thebibliography}

\end{document}